\documentclass[format=acmsmall, review=false, screen=true]{acmart}

\usepackage{booktabs} 

\usepackage{RobStd}
\usepackage{multirow}
\usepackage{algorithm}
\usepackage{float}
\usepackage{algpseudocode}
\usepackage{cellspace}
\usepackage{makecell}
\usepackage{dblfloatfix}
\usepackage{color}

\newcolumntype{"}{@{\hskip\tabcolsep\vrule width 1pt\hskip\tabcolsep}}

\makeatletter
\newenvironment{breakablealgorithm}
  {
   \begin{center}
     \refstepcounter{algorithm}
     \hrule height.8pt depth0pt \kern2pt
     \renewcommand{\caption}[2][\relax]{
       {\raggedright\textbf{\ALG@name~\thealgorithm} ##2\par}%
       \ifx\relax##1\relax 
         \addcontentsline{loa}{algorithm}{\protect\numberline{\thealgorithm}##2}%
       \else 
         \addcontentsline{loa}{algorithm}{\protect\numberline{\thealgorithm}##1}%
       \fi
       \kern2pt\hrule\kern2pt
     }
  }{
     \kern2pt\hrule\relax
   \end{center}
  }
\makeatother

\acmJournal{JETC}

\setcopyright{acmlicensed}

\acmDOI{0000001.0000001}

\begin{document}
\title[A mixed signal architecture for convolutional neural networks]{A mixed signal architecture for convolutional neural networks}

\author{Qiuwen Lou}
\orcid{1234-5678-9012-3456}
\affiliation{%
  \institution{University of Notre Dame}
  \streetaddress{100 Notre Dame Avenue}
  \city{Notre Dame}
  \state{IN}
  \postcode{46637}
  \country{USA}}
\email{qlou@nd.edu}
\author{Chenyun Pan}
\affiliation{%
  \institution{University of Kansas}
  \city{Lawrence}
  \state{Kansas}
  \country{USA}
}
\email{chenyun.pan@gatech.edu}
\author{John McGuinness}
\affiliation{%
 \institution{University of Notre Dame}
 \streetaddress{100 Notre Dame Avenue}
 \city{Notre Dame}
 \state{IN}
 \country{USA}}
\email{john.mcguinness.7@nd.edu}
\author{Andras Horvath}
\affiliation{%
  \institution{Pazmany Peter Catholic University}
  \streetaddress{Szentkiralyi u. 28, 1088}
  \city{Budapest}
  \country{Hungary}
}
\email{horvath.andras@itk.ppke.hu}
\author{Azad Naeemi}
\affiliation{%
  \institution{Georgia Institute of Technology}
  \city{Atlanta}
  \state{Georgia}
  \country{USA}}
\email{azad@gatech.edu}
\author{Michael Niemier}
\author{X. Sharon Hu}
\affiliation{%
  \institution{University of Notre Dame}
  \department{Department of Computer Science and Engineering}
  \city{Notre Dame}
  \state{IN}
  \postcode{46556}
  \country{USA}
}

\begin{abstract}
Deep neural network (DNN) accelerators with improved energy and delay are desirable for meeting the requirements of hardware targeted for IoT and edge computing systems. Convolutional neural networks (CoNNs) belong to one of the most popular types of DNN architectures. This paper presents the design and evaluation of an accelerator for CoNNs. The system-level architecture is based on mixed-signal, cellular neural networks (CeNNs). Specifically, we present (i) the implementation of different layers, including convolution, ReLU, and pooling, in a CoNN using CeNN, (ii) modified CoNN structures with CeNN-friendly layers to reduce computational overheads typically associated with a CoNN, (iii) a mixed-signal CeNN architecture that performs CoNN computations in the analog and mixed signal domain, and (iv) design space exploration that identifies what CeNN-based algorithm and architectural features fare best compared to existing algorithms and architectures when evaluated over common datasets -- MNIST and CIFAR-10. Notably, the proposed approach can lead to 8.7$\times$ improvements in energy-delay product (EDP) per digit classification for the MNIST dataset at iso-accuracy when compared with the state-of-the-art DNN engine, while our approach could offer 4.3$\times$ improvements in EDP when compared to other network implementations for the CIFAR-10 dataset.
\end{abstract}

%
%

\ccsdesc[500]{Hardware~Application specific integrated circuits}
\ccsdesc[300]{Hardware~Application specific processors}

%
%

\keywords{Hardware accelerator, Convolutional neural networks, Analog circuits}

\maketitle

\renewcommand{\shortauthors}{Q. Lou et al.}

\section{Introduction}
\label{sec:1}
In the machine learning community, there is great interest in developing computational models to solve problems related to computer vision \cite{ML_ImageNet_short}, 
speech recognition \cite{ML_Speech}, information security \cite{ML_Security}, 
climate modeling \cite{ML_Climate}, etc. To improve the {\it delay and energy efficiency} of computational tasks related to both inference and training, the hardware design and architecture communities are considering how hardware can best be employed to realize algorithms/models from the machine learning community. 
Approaches include application specific circuits (ASICs) to accelerate deep neural networks (DNNs) \cite{Reagen16_short, 2017_ISSCC_Whatmough_short} and convolutional neural networks (CoNNs) \cite{Moons17_short}, neural processing units (NPUs) \cite{Hashemi17_short}, hardware realizations of spiking neural networks \cite{2015_NIPS_Esser_short, 2015_VLSI_Kim_short}, etc.

When considering application-specific hardware to support neural networks, it is important that said hardware can implement networks that can be extensible to a large class of networks, and solve a large collection of application-level problems. Deep neural networks (DNNs) represent a class of such networks and have demonstrated their strength in applications such as playing the game of Go \cite{alphago}, image and video analysis \cite{ML_ImageNet_short}, target tracking \cite{vot_report}, etc. In this paper, we use convolutional neural network (CoNN) as a case study for DNNs due to its general prevalence. CoNNs are computationally intensive, which could lead to high latency and energy for inference and even higher latency/energy for training. The focus of this paper is on developing a low energy/delay mixed-signal system based on cellular neural networks (CeNNs) for realizing CoNN.

A Cellular Nonlinear/Neural Network (CeNN) is an analog computing architecture~\cite{Chua88} that could be well suited for many information processing tasks.
In a CeNN, identical processing units (called cells) process analog information in a concurrent manner. Interconnection between cells is typically local (i.e., nearest neighbor) and space-invariant.
For spatio-temporal applications, CeNNs can offer vastly superior performance and power efficiency when compared to conventional von Neumann architectures \cite{palitICCAD2015,CeNN_compression}.
Using "CeNNs for CoNN" allows the bulk of the computation associated with a CoNN to be performed in the analog domain. Sensed information could immediately be processed with no analog-to-digital conversion (ADC).  Also, inference-based processing tasks can tolerate lower precision (e.g., Google's TPU employs 8-bit integer matrix multiplies \cite{Google_TPU_short}) typically associated with analog hardware, and can leverage its higher energy efficiency. With this context, we have made the following contributions in this paper.


{\bf (i)} We elaborate the use of CeNN to realize computations that are typically associated with different layers in a CoNN. These layers include convolution, ReLU, and pooling. Based on the implementations for each layer, a baseline CeNN-friendly CoNN for the MNIST problem \cite{lecun2010mnist} is presented\footnote{A preliminary version of the design was presented in \cite{DATE17_short} ($A. Horvath, et al. DATE, 2017$).}. 

{\bf (ii)} We introduce an improved CoNN model for the MNIST problem to support CeNN-friendly layers/algorithms that could ultimately improve figures of merit (FOM) such as delay, energy, and accuracy, etc.
Following the same concept, we also develop a CeNN-friendly CoNN for the CIFAR-10 problem \cite{cifar10}.

{\bf (iii)} We present a complete, mixed-signal architecture to support CeNN-friendly CoNN designs. Besides CeNN cells and SRAM to store weights, the architecture includes analog memory to store intermediate feature map data, and ADC and digital circuits for the FC layer computation. The architecture also supports efficient programming/reprogramming CeNN cells. 

We have conducted detail studies of energy, delay, and accuracy per classification for the MNIST and CIFAR-10 datasets, and compared our networks and architecture with other algorithms and architectures \cite{Reagen16_short, 2017_ISSCC_Whatmough_short, Moons17_short, Hashemi17_short, 2015_NIPS_Esser_short, 2015_VLSI_Kim_short} that address the same problem. For the MNIST dataset, at iso-accuracy, our results demonstrate an 8.7$\times$ improvement in energy-delay product (EDP) when compared with a state-of-the-art accelerator. When compared with another recent analog implementation\cite{analog_dnn_short}, a 10.3$\times$ improvement in EDP is observed. For the CIFAR-10 dataset, a 4.3$\times$ improvement in EDP is observed when comparing with a state-of-the-art quantized approach \cite{Hashemi17_short}. 

The rest of the paper is structured as follows. Sec. \ref{sec:2} gives a general discussion of CeNNs and existing CoNN accelerators.  In Sec. \ref{sec:3}, we present the implementation of CoNN layers in CeNNs. Our baseline network designs as well as other algorithmic changes and network topologies that might be well-suited for our architecture are given in Sec. \ref{sec:4}. Sec. \ref{sec:5} describes our proposed architecture, including CeNN cell design, and simulations of various core architectural components. Evaluation and benchmarking results are presented in Sec. \ref{sec:6}. Lastly, Sec. \ref{sec:7} concludes the paper.

\section{Background}
\label{sec:2}
Here, we briefly review the basic concepts of CeNN and accelerator designs for CoNN.
\subsection{CeNN basics}
\label{ssec:2a}
\begin{figure}[!t]
    \begin{centering}
    \vspace{-0.15in} 
    \includegraphics[width=3.25in]{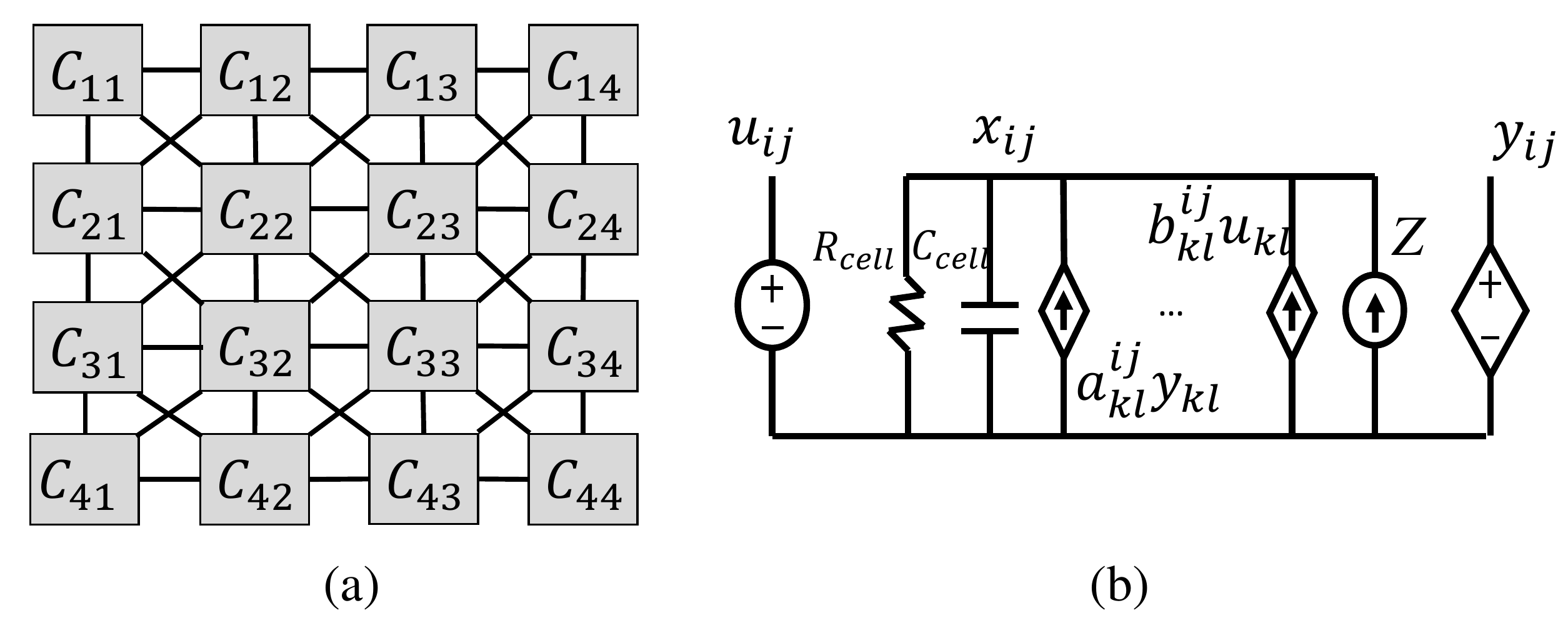}
    \caption{(a) CeNN array architecture; (b) CeNN cell circuitry.}
    \vspace{-0.1in}
    \label{fig:CNN}
    \end{centering}
 \end{figure}

A CeNN architecture is a spatially invariant, $M \times N$ array of identical cells (Fig. \ref{fig:CNN}a) \cite{DATE17_short}. Each cell $C_{ij}$ has identical connections with adjacent cells in a predefined neighborhood. These neighborhood cells are denoted as $N_r(i,j)$ of radius $r$ (i.e., a given cell communicates with other cells within a neighborhood $r$).
The number of cells ($m$) in the neighborhood is given by the expression $m=(2r+1)^2$. 
($r$ is typically 1, which suggests that each cell interacts with only its immediate neighbors.)

A CeNN cell is comprised of one resistor, one capacitor, $2m$ linear voltage controlled current sources (VCCSs), and one fixed current source (Fig. \ref{fig:CNN}b). A cell's input, state, and the output of a given cell $C_{ij}$, corresponds to the nodal voltages, $u_{ij}$, $x_{ij}$, and $y_{ij}$ respectively. VCCSs controlled by input and output voltages of each neighbor deliver feedforward and feedback currents to a given cell. To understand CeNN cell dynamics, we can simply assume a system of $M \times N$ ordinary differential equations.  Each equation is simply the Kirchhoff's Current Law (KCL) at the state nodes of the corresponding cells (Eq.~\ref{eqn_conv_equilibrium}). CeNN cells also employ a non-linear sigmoid-like transfer function at the output (see Eq. \ref{eq:ReLU2}). 

\begin{align}
\vspace{-0.12in}
C_{cell}\dfrac{dx_{ij}\left(t\right)}{dt} &= -\dfrac{x_{ij}\left(t\right)}{R_{cell}}
+ \displaystyle\sum_{C_{kl} \in N_r\left(i,j\right)} a_{ij,kl}y_{kl}\left(t\right) 
+ \displaystyle\sum_{C_{kl} \in N_r\left(i,j\right)} b_{ij,kl}u_{kl} + Z
\label{eqn_conv_equilibrium}
\vspace{-0.1in}
\end{align}

\begin{equation}
  \label{eq:ReLU2}
  y_{k,l}= \frac{1}{2} \left | x_{k,l} + 1\right |- \frac{1}{2} \left | x_{k,l} - 1\right |.
\end{equation}

Feedback and feed-forward weights from cell $C_{kl}$ to cell $C_{ij}$ are captured by the parameters $a_{ij,kl}$ and $b_{ij,kl}$, respectively.  $a_{ij,kl}$, and $b_{ij,kl}$ are space invariant and are denoted by two $(2r+1) \times (2r+1)$ matrices.  (If $r=1$, matrices are $3 \times 3$.)  Matrices of $a$ and $b$ parameters are referred to as templates -- where $A$ and $B$ are the feedback and feed-forward templates, respectively. Template values are the coefficients in the differential equation, and can either be a constant to reflect a linear relationship between cells, or a non-linear function (which can be dependent on the input or state of the corresponding neighboring cell) to reflect non-linear relationship between cells.
Design flexibility is further enhanced by the fixed bias current $Z$. This provides a means to adjust total current flowing into a cell. By selecting values for $A$, $B$, and $Z$, CeNNs can solve a wide range of problems.

Various circuits including inverters, Gilbert multipliers, operational transconductance amplifiers (OTAs), etc can be used as VCCSs in CeNN. \cite{Molinar07,Wang98}. For the work to be discussed in this paper, we assume the OTA design from \cite{Qiuwen15_short}. OTAs provide a large linear range for voltage to current conversion, and can implement a wide range of transconductances that could be used for different CeNN template implementations. Furthermore, these OTAs can also be used to implement {\it Non-linear} templates, which leads to CeNNs with richer functionality \cite{Qiuwen15_short}. 

\vspace{-0.05in}

\subsection{Convolutional neural network accelerators}
Due to the high computational complexity of CoNNs, various hardware platforms are used to enable the efficient processing of DNNs, including GPUs, FPGAs, ASICs, etc. Specifically, there is a growing interest in using ASICs to provide more specialized acceleration of DNN computation. A recent review paper summarized these approaches in \cite{deep_learning_accelerator_survey_short}. Both digital and/or analog circuitries are proposed to implement these accelerators. In the digital domain, typical approaches include using optimized dataflow to efficiently reduce the data movement overhead for the dense matrix multiplication operation \cite{eyeriss_short} , or implementing sparse matrix multiplication by applying pruning to the network \cite{eie_short}. Recently, analog implementations have also been proposed to accelerate deep learning processes. Work in \cite{analog_dnn_short} embedded a charge sharing scheme into SRAM cells to reduce the overhead of memory accesses. Work in \cite{isaac} uses a crossbar circuit with memristors to speed up the inference of deep neural networks, 

\section{CeNN implementation of CoNN computations}
\label{sec:3}

As pointed out earlier, CeNNs have a number of benefits such as {\bf (i)} ease of implementation in VLSI, {\bf (ii)} low energy due to its nature fit for analog realization, {\bf (iii)} Turing complete, etc.    We show in this section that all the core operations in a CoNN can be readily implemented with CeNNs. 
In a CoNN, every layer typically implements a simple operation that might include:  {\bf (i)} convolutions, {\bf (ii)} non-linear operations (usually a rectifier), {\bf (iii)} pooling operations, and {\bf (iv)} fully connected layers.
Below, we describe how each of these layers can map to a CeNN.  A more detailed description of the operations and how the layered network itself can be built can be found in \cite{LeCunDeep}\cite{GoogleNet}. We will also discuss our network design in Sec. \ref{sec:4}.

\subsection{Convolution}
\label{sec:3:Convolution}
Convolution layers are used to detect and extract different feature maps on input data as the summation of the point-wise multiplication of the feature map and the convolutional kernel. One map is the input image ($f$), and the convolutional kernel encodes a desired feature ($g$) to be detected by some operation. It is easy to see that a convolution has the highest response at positions where the desired feature appears. The convolution operation can be defined per Eq. \ref{eq:Convolution1}. The exact convolutional kernels are optimized during training. 

\begin{equation}
  \label{eq:Convolution1}
  f * g (i,j) = \sum_{k,l=-\infty}^{\infty}f(i-k,j-l)g(k,l) \vspace*{-1ex}
\end{equation}

As can be seen from Eq. \ref{eqn_conv_equilibrium}, with the application of the feed-forward template (denoted as $b_{ij,kl}$), one CeNN can implement a convolutional kernel for a feature map in a straightforward manner. 
Then, all these feature maps after convolutional operations need to sum up together.
We will discuss the mechanism for achieving this in Sec. \ref{sec:5}.

Due to the sigmoid function within the CeNN equation, the output of CeNN is thresholded to the range (-1, 1). However, in the CoNN computation, the output could be larger than 1 or less than -1, which leads to an error in data representation. However, our initial simulation results suggest that this error does not impact the overall classification accuracy in the networks considered in this paper.

\subsection{Rectified Linear Units}
\label{sec:3:RectifiedLinearUnits}
As CoNNs are built and designed for recognition purposes and classification tasks, non-linear operations are required. 
Perhaps the most commonly used non-linearity in deep learning architectures \cite{DahlReLU} is the rectified linear unit (ReLU) that per Eq. \ref{eq:ReLU1}, thresholds every value below zero.

\begin{equation}
  \label{eq:ReLU1}
  R(x)={ 
    \begin{Bmatrix}
        0, if  x\leq 0\\ 
        x, if  x>0
    \end{Bmatrix} 
    }
\end{equation}

In a CeNN, the ReLU operation can be implemented using a non-linear template. In CeNN theory, nonlinear templates are usually noted as $\hat{D}$ templates in parallel with $A$ templates and $B$ templates. To realize the required non-linear computation here, one can define an additional template implementing the non-linear function of the ReLU operation: $\hat{D}(x_{i,j})= \text{ max}(0, x_{i,j})$. This function sets all negative values to zero and leaves the positive values unchanged, hence it directly implements Eq. \ref{eq:ReLU1}.
That said, {\bf (i)} while non-linear templates are well established in the {\it theory} of CeNNs, {\bf (ii)} the application of a non-linear function has obvious computational utility, and {\bf (iii)} non-linear templates can be easily simulated, in practice, non-linear operations are much more difficult to realize. While existing hardware considers non-linear template implementations \cite{Qiuwen15_short}, it may still not exactly mimic the behavior of non-linear templates. (We will discuss this in more detail in Sec. \ref{sec:3:non_linear}.)

Alternatively, as the CeNN-UM is Turing complete, all non-linear templates can be implemented as a series of linear templates together with the implicit CeNN non-linearity (i.e. sigmoid output, see Eq. \ref{eq:ReLU2}) \cite{CNNUM}. This implicit CeNN non-linearity is widely implemented in real devices such as the ACE16k chip \cite{ace16k} or the SPS 02 Smart Photosensor from Toshiba \cite{SPS02}.
In the CoNN case, the ReLU operation can be rewritten as a series of linear operations (with the implicit CeNN non-linearity) by applying templates below.

First, one can execute the feed-forward template given by Eq. \ref{eq:ReLU3}, which simply decreases all values by $1$. Because the standard CeNN non-linearity thresholds all values in a CeNN array below $-1$, after this shift all values between $-1$ and $0$ are simply set to $-1$.

\begin{equation}
  \label{eq:ReLU3}
  B_1 = 
    \begin{bmatrix}
      0 & 0 & 0 \\ 
      0 & 1 & 0\\ 
      0 & 0 & 0
    \end{bmatrix},
  Z=-1
\end{equation}
Next, one can shift the values back, (i.e., increase them by 1) by applying the template operation in Eq. \ref{eq:ReLU4}:

\begin{equation}
\label{eq:ReLU4}
  B_2 = 
    \begin{bmatrix}
      0 & 0 & 0 \\ 
      0 & 1 & 0 \\ 
      0 & 0 & 0
    \end{bmatrix},
  Z=1
\end{equation}

As the non-linearity thresholds a given value, these two linear operations implement the required ReLU operator, i.e., leaving all positive values unchanged, and thresholds all values below 0. 


\subsection{Pooling}
\label{sec:3:Pooling}
Pooling operations are employed to decrease the amount of information flow between consecutive layers in a deep neural network to compensate for the effects of small translations. Two pooling approaches are widely used in CoNN -- max pooling and average pooling. Here, we discuss the implementations of both pooling approaches using CeNN.

\subsubsection{Max pooling}
\label{sec:3:max_pooling}
A max pooling operation selects the maximum element in a region around every value per Eq. \ref{eq:Pooling1}:
\begin{equation}
  \label{eq:Pooling1}
  P(i,j) = \text{max}_{k,l \in S}f(i-k,j-l)
\end{equation}

\noindent Similar to ReLU, max pooling is also a non-linear function. As before, functionality associated with max pooling can also be realized with a sequence of {\it linear} operations. We use a pooling operation with a 3$\times$3 receptive field as an example to illustrate the process. The idea here is to compare the intensity of each pixel in the image with all its neighbors in succession (with a radius of 1 in the 3$\times$3 case). We use $x_{i,j}$ to represent the intensity for pixel $(i,j)$. For each comparison, if the intensity of its neighbor pixel (defined as $x_{k,l}$) is larger than $x_{i,j}$, we use $x_{k,l}$ to replace $x_{i,j}$ in the location $(i,j)$, otherwise $x_{i,j}$ remains unchanged. By making comparisons with all neighboring pixels, the value of $x_{i,j}$ can be set to the magnitude of all of its neighbors. 

We developed a sequence of CeNN templates to realize the comparison between $x_{i,j}$ and all its neighboring pixels, $x_{k,l}$. Then, by simply rotating the templates, we can easily compare $x_{i,j}$ to other neighbor pixels.
Downsampling could be performed afterwards to extract the maximum value within a certain range if needed.
The detailed CeNN operations to realize the comparison can be broken down into 4 steps, and are summarized as follows.
{\bf (i)} Apply the linear DIFF template shown in Eq. \ref{eq:Pooling3}:
\begin{equation}
  \label{eq:Pooling3}
  B_1 = 
    \begin{bmatrix}
      0 & 0.5 & 0 \\ 
      0 & -0.5  & 0\\ 
      0 & 0  & 0
    \end{bmatrix},     
  Z = -1
\end{equation}
The output after applying this template is $y = -0.5x_{i,j} + 0.5x_{k,l}-1$. After applying the sigmoid function, $y = -1$ if $x_{i,j}\geq x_{k,l}$, otherwise $y$ remains unchanged.
{\bf (ii)} Apply the linear INC template in Eq. \ref{eq:Pooling4} to shift the pixel intensity up. After this operation, $y$ becomes 0 if $x_{i,j}\geq x_{k,l}$, otherwise $y = -0.5x_{i,j} + 0.5x_{k,l}$.

\begin{equation}
  \label{eq:Pooling4}
  B_2 = 
    \begin{bmatrix}
      0 & 0  & 0 \\ 
      0 & 1  & 0\\ 
      0 & 0  & 0
    \end{bmatrix},
  Z = 1
\end{equation}

\noindent {\bf (iii)} Apply the CeNN MULT template to multiply $y$ by 2. Thus, $y = 0$ if $x_{i,j}\geq x_{k,l}$, otherwise $y = -x_{i,j} + x_{k,l}$. {\bf (iv)} Add $x_{i,j}$ 
to $y$ to obtain the maximum between $x_{k,l}$ and $x_{i,j}$, and use it to update the intensity in the location $(i,j)$.

\subsubsection{Average pooling}
Per Sec. \ref{sec:3:max_pooling}, a max pooling operation with linear CeNN templates requires up to 16 computational steps. (Each comparison requires 4 steps, while the pixel needs to  compare with (at least) its neighboring 4 pixels.) That said, average pooling can be used in lieu of max pooling, and may have only a nominal impact on the classification accuracy in certain scenarios \cite{ave_pooling_short}. Average pooling operations can be easily realized with CeNNs -- in fact, only one template operation is required. To perform an average pooling operation in $2 \times 2$ or $3\times 3$ grids, one can simply employ the $B$ templates in Eq. \ref{eqn:avg_pooling} ($Z=0$). 
\begin{equation}
  \label{eqn:avg_pooling}
  B_{2x2} = 
    \begin{bmatrix}
      1/4 & 1/4 & 0 \\ 
      1/4 & 1/4 & 0\\ 
      0   & 0   & 0
    \end{bmatrix},
  \;\;
  B_{3x3} = 
    \begin{bmatrix}
      1/9 & 1/9 & 1/9 \\ 
      1/9 & 1/9 & 1/9 \\ 
      1/9 & 1/9 & 1/9
    \end{bmatrix}
\end{equation}



%
%

\subsection{Non-linear template operations}
\label{sec:3:non_linear}
While CeNN templates typically suggest linear relationships between cells, non-linear relationships are also possible and can be highly beneficial. (As noted earlier, while non-linear template operations are well-supported by CeNN theory, in hardware realizations, linear operations are more common owing to the complexity of the circuitry required for non-linear steps.)  That said, we also consider what impact non-linear template operations may ultimately have on application-level FOM.


We consider non-linear implementations of ReLU and pooling per \cite{Qiuwen15_short}. The non-linear OTA based I-V characteristic shown in \cite{Qiuwen15_short} can directly mimic the ReLU function discussed in Sec. \ref{sec:3:RectifiedLinearUnits}. 
The pooling operation can also be implemented by the non-linear, GLOBMAX template, which can be found in the standard CeNN template library\cite{temlib}. The GLOBMAX operation selects the maximum value in the neighborhood of a cell in a CeNN array and propagates it through the array. By setting the execution time of the template accordingly, one can easily set how far the maximum values can propagate/which regions the maximum values can fill. Here, the non-linear templates can also be implemented by using the $\hat{D}$ type non-linear function as given in Eq. \ref{eq:Pooling2}.

\begin{equation}
  \label{eq:Pooling2}
  \hat{D}(x_{i,j})= { 
     \begin{Bmatrix}
       -\frac{1}{8}x, \text{ if } x\leq 0\\ 
       0, \text{ if } x>0
     \end{Bmatrix} 
     }
\end{equation}


\subsection{Fully-Connected Layers}
\label{sec:3:FullyConnectedLayers}
The operations described above are used in local feature extractors and can extract complex feature maps from a given input image. However, to accomplish classification, one must convert said feature maps into a scalar index value associated with the selected class.  While various machine learning algorithms (e.g., SVMs) can be used for this, a common approach is to employ a fully connected (FC) layer and associated neurons. The FC layer considers information globally and unifies local features from the lower layers. It can be defined as a pixel-wise dot product between a weight map and the feature map. This product can be used as a classification result, which captures how strongly the data belongs to a class and the product is calculated for every class independently. The index of the largest classification result can be selected and associated with the input data. 

CeNNs can be readily used to implement the dot product function in the FC layer. However, if for large feature maps and weight maps, i.e., the point-wise calculation for vector length over 9. CeNN would require large $r's$, hence cannot efficiently implement such FC layers. To overcome this challenge, one can leverage a digital processor (e.g., per \cite{7298274}) to perform the FC layer function.


-+ \section{CeNN-based CoNNs for two case studies}
\label{sec:4}

As mentioned in the previous section, {\bf (a)} CeNNs could operate in the analog domain -- which could result in lower power consumption/improved energy efficiency \cite{LowPowerCNN}, and {\bf (b)} CeNNs are Turing complete \cite{Roska} and could provide a richer library of functionality than which is typically associated with CoNNs. In this section, we consider how the topographic, highly parallel CeNN architecture can efficiently implement deep-learning operations/CoNNs.

CeNNs are typically comprised of a single layer of processing elements (PEs).  Thus, while most CeNN hardware implementations lack the layered structure of CoNNs, by using local memory and CeNN reprogramming (commonly available on every realized CeNN chip \cite{ace16k} as will be discussed), a cascade of said operations can be realized by re-using the result of each previous processing layer \cite{Roska}.  
One could also simply use multiple CeNNs to compute different feature maps in each layer in parallel. These CeNNs need to communicate with each other, e.g., in order to sum values for different feature maps. Below, we show how the layered CoNNs can be realized with layers of CeNNs through two case studies: {\bf (i) MNIST}, {\bf (ii) CIFAR-10}.

\subsection{CeNN-based CoNNs for MNIST}
\label{sec:4:MNIST}
Using the building blocks described above, we have developed several CeNN-friendly structures for the MNIST problem. In the MNIST handwritten digit classification task \cite{MNIST}, a system must analyze and classify what digit (0-9) is represented by a 28 $\times$ 28 pixel gray scale image. There are 60,000 images in the training set, and 10,000 images in the test set.

To develop the CeNN-friendly CoNN, we leverage the following two observations. First, all computational kernels are best to be restricted to a CeNN friendly size of $3 \times 3$.  In some sense, this could be viewed as a "departure" from larger kernel sizes (e.g., $7 \times 7$ or larger) that may be common in CoNNs.  It should be noted that larger kernels are acceptable according to the CeNN theory (i.e., per Sec. \ref{sec:2}, a neighborhood's radius $r$ could easily be larger than 1).  However, due to increased connectivity requirements, said kernels are infrequently realized in hardware.  That said, the $3 \times 3$ kernel size is not necessarily a restriction. Recent works \cite{vgg-net} suggests that larger kernels can be estimated by using a series of $3 \times 3$ kernels with fewer parameters. Again, this maps well to CeNN hardware. Second, per the discussion in Sec. \ref{sec:3}, all template operations for the convolution, ReLU, and pooling steps are feed-forward (B) templates. The feedback template (A) is not used in any of the feature extracting operations (i.e., per Eq. \ref{eqn_conv_equilibrium}, all values would simply be 0).

During network development, we use TensorFlow to train the network with full precision to obtain accuracy data. We use stochastic gradient descent for training, with the initial learning rate set to $10^{-2}$. We have also implemented a more versatile/adjustable training framework in MATLAB. The MATLAB based simulator extracts weights from the trained model (from TensorFlow), and performs inference in conjunction with CeNN operations at the precision that is equivalent to actual hardware. Our network learns the parameters of the B-type templates for the convolution kernels.  (Per Sec. \ref{sec:3}, the B-template values for the ReLU and pooling layers are fixed.) 

\begin{figure}[!t]
  \centering
  \includegraphics[width=5in]{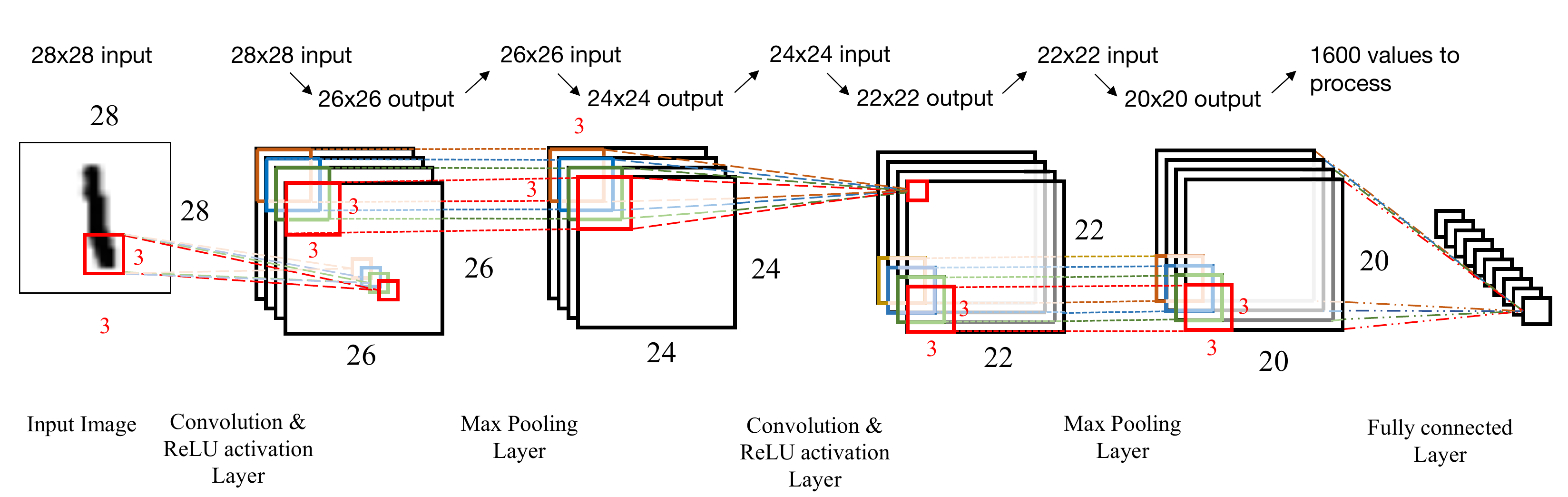}
  \vspace{-3ex}
  \caption{CeNN-friendly CoNN for the MNIST problem -- design 1.}
  \label{fig:design1}
  \vspace{-0.2in}
\end{figure}

Following the observations and process described above, we develop a layered, CeNN-friendly network to solve the MNIST problem. The network topology is shown in Fig. \ref{fig:design1}. The network contains two convolution layers, and each layer contains 4 feature maps. There is also an FC layer that follows the two convolution layers to obtain the classification results. The baseline network is designed using maximum pooling and linear templates to potentially maximize the classification accuracy. However, we also study the network accuracy for average pooling and alternatives based on non-linear templates, to evaluate tradeoffs in terms of accuracy, delay, energy, etc. to be discussed. 

The accuracy for different design options for the network are summarized in the second column in Table \ref{tbl:mnist_accuracy}. From the table, we can see that max pooling generally leads to better accuracy than average pooling. 
The non-linear template implementation is also less accurate than the linear implementation for max pooling. This is mainly because the GLOBALMAX template is an approximation for the max pooling, and it is {\textit not} as accurate as the linear template approach. 

\begin{table}[b]
 \caption{Classification accuracy for different CoNN designs for the MNIST problem}
  \vspace{-0.05in}
 \centering
 \scriptsize
\renewcommand{\arraystretch}{1.35}
 \begin{tabular}{| c |c |c | }
 \hline
   Approach & Network in Fig. \ref{fig:design1} & Network in Fig. \ref{fig:design2}\\
  \hline
  Baseline & 98.1\% & 97.8\%\\
 \hline
   Average pooling & 97.5\% & 96.7\% \\
 \hline
   Nonlinear templates & 93.1\%& 91.5\% \\
  \hline
  \end{tabular}
  \label{tbl:mnist_accuracy}
  \vspace{-0.1in}
\end{table}

\vspace{-3pt}

\subsection{Eliminating FC layers}
\label{sec:4:EliminatingFCLayer}
One of the potential challenges of a network with a fully-connected layer shown in Fig. \ref{fig:design1} is the need to convert analog CeNN data into a digital representation to perform computations associated with an FC layer since an FC layer is not CeNN friendly (see Sec. \ref{sec:3:FullyConnectedLayers}).
To reduce the impact of analog-to-digital conversion and associated FC layer computation, we have designed an alternative network for MNIST digit classification to perform computations associated with an FC layer. 

\begin{figure}
  \centering
  \includegraphics[width=5in]{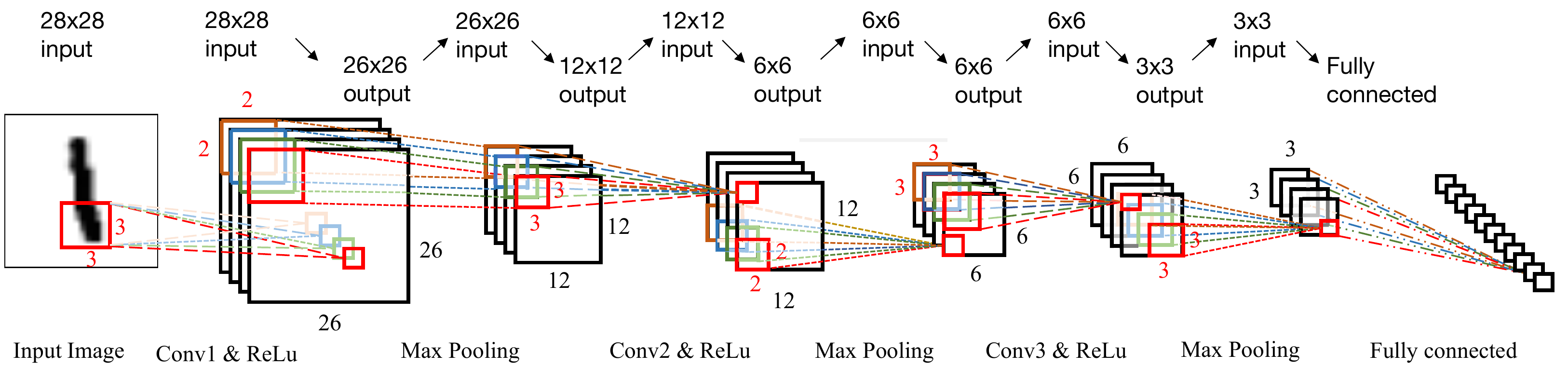}
  \vspace{-3ex}
  \caption{CeNN-friendly CoNN for the MNIST problem -- design 2.}
  \label{fig:design2}
  \vspace{-2ex}
\end{figure}

In this alternative network (Fig. \ref{fig:design2}), the weights (and image sizes) associated with the last layer of the network are reduced to CeNN-friendly, $3 \times 3 $ kernels. Changes include modifications to the pooling layer. In the network in Fig. \ref{fig:design1}, max pooling is achieved by propagating the maximum pixel value to all neighbors within a certain region specified by the network design. However, the sizes of these feature maps do not change.  For the network in Fig. \ref{fig:design2}, the maximum value is propagated within a $2 \times 2$ grid to form a group, and only one maximum pixel value in each group is extracted to be processed in the next stage of the network. Thus, the network size is reduced by a factor of two with each pooling layer. For the implementation of downsampling through max pooling, after a pooling operation is completed, for each a $2 \times 2$ grid within a feature map, only one pixel is required to write to an analog memory array for the next stage processing. 
In the network in Fig. \ref{fig:design2}, three pooling layers are required to properly downsize an image and obtain reasonable accuracy.  
The final computational steps associated with this alternative network are readily amenable for CeNN hardware implementations. However, both the image size and the kernel size are reduced to $3 \times 3$.

Potential overheads associated with FC layer computations are reduced as only the final results (10 probability values corresponding to the number of image classes) must be sent to any digital logic and/or CPU (in lieu of the 16,000 signals associated with the network in Fig. \ref{fig:design1}). Downsampling may also impact classification energy, as smaller subsets of the CeNN array can be used for computations associated with successive layers in the network. Again, we evaluate the accuracy of this proposed approach by using average pooling, nonlinear templates, etc. The results are shown in the third column in Table \ref{tbl:mnist_accuracy}. 
In general, these accuracy numbers are still close to the baseline design discussed in Sec. \ref{sec:4:MNIST}.

In general, this strategy should be applicable to any network, regardless of its depth, width and the kernel sizes employed. By properly downsampling the feature map in the relevant layer (i.e., to reduce the feature map size by 1/2 or 1/3 when needed), we can eventually obtain a 3x3 feature map in the last layer of a given network.

\subsection{CeNN-based CoNNs for CIFAR-10}

\label{sec:4:cifar10}
The networks proposed in Sec. \ref{sec:4:MNIST} and Sec. \ref{sec:4:EliminatingFCLayer} for MNIST are relatively simple compared with state-of-the-art networks. Typically, to solve more complex problem, larger networks with more layers/feature maps are required. 
In this subsection, we discuss our design for larger CeNN-friendly CoNNs.

\begin{figure}[!t]
  \centering
  \includegraphics[width=5in]{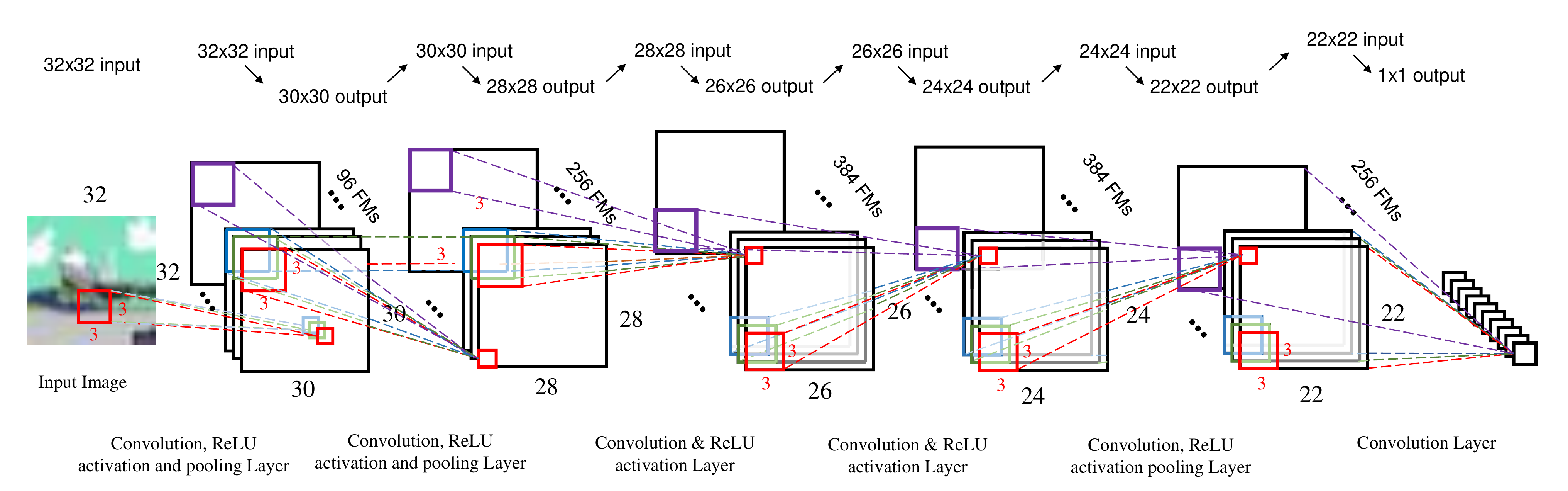}
  \vspace{-3ex}
  \caption{CeNN-friendly CoNN for CIFAR-10 problem.}
  \label{fig:cifar_10}
  \vspace{-0.15in}
\end{figure}
As a case study, we use CIFAR-10 as the dataset, which consists of 50,000 images in the training set, 10,000 images in the validation set and 10,000 images in the test set. These images are all color images with RGB channel. There are 10 classes with different objects (e.g., airplane, automobile, bird, etc.) within the dataset.
Each image belongs to one class, with a size of 32 $\times$ 32. During the inference stage, the network must predict which class the image belongs to. 

We use modified AlexNet \cite{ML_ImageNet_short} network to solve the CIFAR-10 problem.
AlexNet is originally used to solve ImageNet \cite{imagenet}, which is a more complex problem. Thus, we expect our modification still leads to reasonable accuracy for CIFAR-10. We perform our modifications on AlexNet to {\bf (i)} enable the modified network to solve the CIFAR-10 problem, and {\bf (ii)} make the network CeNN-friendly.
Specifically, our main modifications are summarized as follows: {\bf (i)} For all convolution layers in AlexNet, the kernel sizes are changed to 3 $\times$ 3 so that it is readily amendable to CeNNs with the same template size. {\bf (ii)} We remove the FC layer in the AlexNet since it is not CeNN-friendly, and use a convolution layer with 10 outputs as the last layer, to obtain the classification probabilities. {\bf (iii)} Downsampling in the pooling layer is not used in the modified network in order to retain reasonable model size. The network architecture is shown in Fig. \ref{fig:cifar_10}.

We use the network in Fig. \ref{fig:cifar_10} as a baseline, and explore the design space by (1) changing the number of feature maps in each layer, (2) using the downsampling approach mentioned in Sec. \ref{sec:4:EliminatingFCLayer}.

In the baseline, the feature maps for the first 5 convolution layers are the same as AlexNet (C96-C256-C384-C384-C256). We also considered feature map sizes of C64-C128-C256-C256-C128 and C64-C128-C128-C128-C64. We use the Adam algorithm \cite{adam} to train the network, with learning rate set to $10^{-4}$. The accuracy data for different design options are summarized in Table \ref{tbl:cifar_accuracy}. The accuracies only drop for 1.6\% and 2.17\% with the decrease of the network size. 
Therefore, we also consider these two networks in the benchmarking efforts discussed in Sec. \ref{sec:6}. 

We also use the approach mentioned in Sec. \ref{sec:4:EliminatingFCLayer} to resize the feature maps of selective layers, to make the size of each feature map in last layer 3$\times$3. The feature maps of the five layers in the CeNN-friendly AlexNet become 32x32-->16x16-->8x8-->4x4-->3x3, which makes the last FC layer CeNN-friendly. The accuracy of the network with this downsampling strategy reaches 80.5\%. Since this approach does not give as good accuracy as these approaches that change the size of feature map discussed above, we do not include it in the benchmarking effort discussed in Sec. \ref{sec:6}.

\begin{table}[!b]
 \vspace{-0.2in}
 \caption{Classification accuracy for different CoNN designs for the CIFAR-10 problem}
  \vspace{-0.1in}
 \centering
 \scriptsize
\renewcommand{\arraystretch}{1.35}
 \begin{tabular}{| c |c |c |c| }
 \hline
   Approach & CeNN-friendly AlexNet &
   CeNN-friendly AlexNet &
   CeNN-friendly AlexNet \\
   & C96-C256-C384-C384-C256 &
   C64-C128-C256-C256-C128 &
   C64-C128-C128-C128-C64\\
  \hline
  Accuracy & 84.5\% & 82.9\% & 81.8\%\\
  \hline
  \end{tabular}
  \label{tbl:cifar_accuracy}
\end{table}

\section{CeNN architectures}
\label{sec:5}
In this section, we introduce our CeNN-based architecture for realizing CeNN-friendly CoNNs. Our architecture is general and programmable for any CoNN that contains convolution, ReLU and pooling layers.
Meanwhile, by changing the configurations (e.g., SRAM size, number of OTAs) and parameters of the circuits (e.g., bias current), our CeNN architecture design could be used to satisfy different precision requirements for the network. Thus, we can explore tradeoffs between accuracy, delay and energy efficiency within the same network. 
We first present our CoNN-based architecture in Sec. \ref{ssec:architecture}. We then describe each component in the architecture, i.e., CeNN cells in Sec. \ref{ssec:cell}, analog memories in Sec. \ref{ssec:analog_momory}, and SRAM in Sec. \ref{sec:5:sram}. We also highlight the dataflow for the CoNN network computation using CeNN architecture.
In Sec. \ref{ssec:fc_layer}, we discuss the need for ADCs and digital circuitry to support computations in an FC layer (i.e., to support networks as discussed in Sec. \ref{sec:4:MNIST}). Finally, we discuss the programming mechanism for the CeNN templates of the architecture. Throughout we also highlight differences between CeNN cell designs presented here as compared to previous work (e.g., \cite{DATE17_short}).

\subsection{Architecture}
\label{ssec:architecture}

\begin{figure}[!t]
    \centering
    \includegraphics[width=5in]{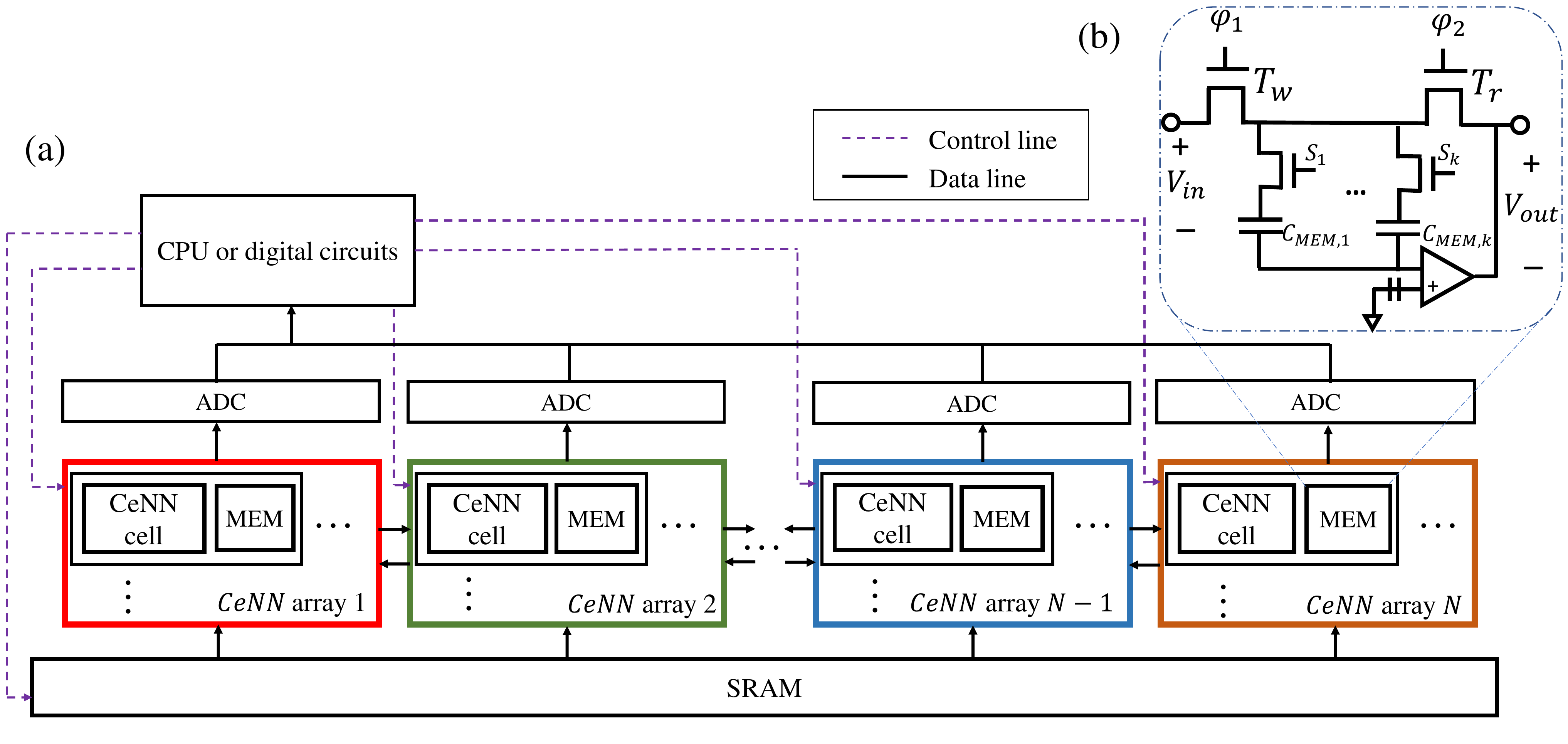}
    \vspace{-8pt}
    \caption{(a) CeNN-based architecture for CoNN operations; (b) Analog memory cell schematic.}
    \label{fig:architecture}
    \vspace{-8pt}
\end{figure}

Our CeNN architecture for (Fig. \ref{fig:architecture}) CoNN computation consists of multiple CeNN arrays (boxes labeled by $CeNN$ array $i$). These arrays are the key components for implementing convolution, ReLU and pooling operations in a CoNN. 
Within each array, there are multiple cells per Sec. \ref{ssec:2a}. The array size can usually accommodate all the image pixels to enable parallel processing of a whole image (extra cells will be power gated to save power). For large images, time multiplexing is used to sequentially process part of the image. The connections between these cells follow the typical CeNN array design as described in Sec. \ref{ssec:2a}.
An SRAM array (the rectangle at the bottom of Fig. \ref{fig:architecture}) is used to store the templates needed for the CeNN computation. How to configure the CeNN templates with the SRAM data is discussed in Sec. \ref{sec:5:sram}. An analog memory array (boxes labeled by MEM) is embedded into each CeNN cell. The analog memory array is used to store intermediate results for the CeNN computation. Each CeNN array is associated with an ADC. The output of the ADC connects to the host processor or a digital logic, which supports computations for FC layers. 

Each CeNN array performs computations associated with one feature map at one time. Thus, $N$ feature maps could perform computations simultaneously with $N$ CeNN arrays. 
Generally in a state-of-the-art CoNN design, there may be hundreds of feature maps. However, it is not possible to accommodate hundreds of CeNNs in a chip due to area and power restrictions. Therefore, these CeNNs need to be time multiplexed to compute different feature maps in one layer, and the intermediate data needs to be stored in the associated analog memory for processing in the next layer. Thus, the number of CeNN arrays should be chosen to balance the power/area of the chip and the degree of parallel computation of feature maps (FMs) in any given layer. 

We use a convolution layer as an example to illustrate how the computation is performed since it is typically the most time/energy consuming layer in state-of-the-art CoNN designs. We assume layer $L_l$ is a convolution layer, and the layer has $C_{l-1}$ feature maps as inputs and $C_{l}$ feature maps as outputs. We assume the number of CeNN arrays is $N$. For each output feature map $FM(l,i)$ in layer $L_l$, the computation required is shown in Eq. \ref{eq:fm_compute}. Namely, each feature map $j$ ($j$ from 1 to $C_{l-1}$) in layer $L_{l-1}$ must convolve with a kernel $K(l,j,i)$, and the sum of the convolution results need to be computed. That is,
\begin{equation}
\label{eq:fm_compute}
FM(l,i) = \sum_{j=1}^{C_{l-1}} K(l,j,i)*FM(l-1,j)
\end{equation}
The computation in Eq. \ref{eq:fm_compute} needs to be repeated $C_l$ times to obtain the results for all the feature maps in Layer $l$.

To compute feature map $FM(l,i)$, we first perform convolution operations on $N$ feature maps in layer $l-1$ from $FM(l-1,1)$ to $FM(l-1,N)$, to obtain $FM_{temp}^{(1)}$ to $FM_{temp}^{(N)}$ (i.e., $FM_{temp}^{(N)} = K(l,N,i)*FM(l-1,N)$). Then we perform $FM_{pSum}^{(1)} = \sum_{i=1}^{N} FM_{temp}^{(i)}$ by leveraging the connections among these CeNNs. The intermediate results $FM_{pSum}^{(1)}$ are stored in the analog memories associated with the CeNN array 1. Similarly, the convolution operation on another $N$ feature maps in layer $l-1$ ($FM(l-1,N+1)$ to $FM(l-1,2N)$) are performed. Again, we compute $FM_{pSum}^{(2)}$ and store it in the analog memories associated with CeNN array 2. We repeat the above process until all the input feature maps convolved with a convolution kernel, and their partial sums (from $FM_{pSum}^{(1)}$ to $FM_{pSum}^{(M)}$, where $M=C_l/N$) are all stored in the analog memories associated with $CeNN_1$ to $CeNN_M$. If the number of CeNNs, $N$, is less than $M$, one CeNN would store more than one feature maps. Then, we sum these partial sums up to obtain the feature map $i$ in layer $L_l$ (i.e., $FM(l,i) = \sum_{q=1}^{M}FM_{pSum}^{(q)}$). Again, the above process is repeated $C_l$ times to obtain all feature maps in layer $L_l$.
The detailed algorithm is shown in Algorithm \ref{alg:dataflow}.
Other type of CoNN layer computations are also summarized in the Algorithm \ref{alg:dataflow}. 
By iteratively using the CeNN architecture, we realize different functionalities.
The relation between the processing time and number of CeNNs for a convolutional layer $l$ can be calculated as in Eq. \ref{eq:delay}.

\begin{equation}
\label{eq:delay}
t = \sum_{l=1}^{l=L} [(\frac{C_lC_{l-1}}{N-1}+\frac{C_lC_{l-1}}{N})(t_{CeNN}+t_{prog}) + \frac{C_lC_{l-1}}{2(N-1)}t_{MEM-read} + \frac{C_lC_{l-1}}{2(N-1)}t_{MEM-write}]
\end{equation}

Here, $t_{CeNN}$ refers to the settling time of an CeNN array, and $t_{MEM-read}$ and $t_{MEM-write}$ are the analog memory read and write time, respectively. $t_{prog}$ refers to the reprogramming time of CeNN (i.e., loading new templates).

In our architecture, the reprogramming or reconfiguration overhead mainly includes reading the bit cells from the SRAM block, and using these outputs to control the switches that power gate OTAs to realize different weight values. The overhead of reading bit cell from the SRAM block dominates. The delay and energy of reading data from the SRAM is accounted for in the evaluation section.
\vspace{0.07in}
\begin{breakablealgorithm}
\label{alg:dataflow}
\caption{CoNN layer computation with CeNN}
\begin{algorithmic}[1]
\Procedure {CeNNforCoNN}
{$K$, $FM(l-1,j), \forall j \in \{0,1, ..., C_{l-1}-1\}$), $L_l$}\\
\Comment{$K$ are template values in the layer, 
 $FM(l-1,j), (\forall j \in \{0,1, ..., C_{l-1}-1\})$ are feature maps from the last layer,
 $L_l$ is the type of layer $l$
 }\;
     
     \If{layer $L_l$ = CONV}\Comment{perform computations in convolution layers}
      \For{i=0 to $C_l-1$}\Comment{compute each feature map $FM(l,i)$ in layer $L_l$}
           \For {q=0 to $\frac{C_{l-1}}{N}-1$}\Comment{compute convolution on all feature maps in layer $L_{l-1}$}
              \For{j=0 to $N-1$}\\
             \Comment{multiplications processed in parallel, summations processed in series}
                    \State $FM_{pSum}^{(q)} = \sum_{j=1}^{N} K(l,q*N+j,i)*FM(l-1,q*N+j)$)
              \EndFor
           \EndFor
        \State $FM(l,i) = \sum_{j=1}^{C_{l-1}} FM_{pSum}^{(q)}$
      \EndFor
         
      \EndIf  
     
     \If{layer $L_l$ = ReLU}
        \Comment{compute ReLU on all feature maps}
        \For{q=0 to $\frac{C_{l-1}}{N}-1$}
           \For{j=0 to $N-1$} \\
           \Comment{$N$ FMs are processed in parallel, steps in ReLU are performed in series} 
            \State $Intermediate(i+q*N) = K(SHIFTLOW)*FM(l-1,j+q*N)$
            
            \State $FM(l,j+q*N) = K(SHIFTBACK)*Intermediate(i+q*N)$
            \EndFor
        \EndFor
      \EndIf
    
     \If{layer $L_l$ = Pooling}
     \Comment{compute pooling on all feature maps}
     \For{q=0 to $\frac{C_{l-1}}{N}-1$}
        \For{j=0 to $N-1$}
        \For{p=0 to 3} \Comment{for each neighbor of the current pixel (see Sec. \ref{sec:3:Pooling}})
         \State $DIFF(p) = K(DIFF(p))* FM(l-1,j)$
         \State $Increase(p) = K(INC)*DIFF(p)$
         \State $Mult(p) = K(MULT)*Increase(p)$
         \State $FM(l,j) = FM(l-1,j)+Mult(p)$
         \EndFor
        \EndFor
       \EndFor
     \EndIf
\EndProcedure
\end{algorithmic}
\end{breakablealgorithm}

The templates of each CeNN can be programmed to implement different kernels in a given CoNN. 
Before each CeNN operation, all the OTAs must be reconfigured to implement different templates. These templates are read from the SRAM block, where all template values are stored. The bitline outputs of the SRAM are connected to the switches of the OTAs. After configuration, CeNN operations are performed. Below, we discuss the key blocks in the CeNN architecture.

\subsection{CeNN Cells design}
\label{ssec:cell}
CeNN arrays are the core computational elements in our architecture. 
The CeNN template values for different layers are determined during the network design phase. For convolutional layers, the templates are the same as weights, which are \textit{trained} by deep neural network frameworks. The templates for ReLU and pooling are discussed in Sec. \ref{sec:3}, and they are independent of the specific problem instance. These template values are read from the SRAM to configure the VCCSs in the CeNN cells. Note that all the cells in an array share the same template values. However, different CeNN arrays may employ the same templates (i.e., for ReLU and pooling layers), or employ different templates (i.e., for convolution layers). 

\begin{figure}[!b]
    \centering
    \vspace{-0.1in}
    \includegraphics[width=4.5in]{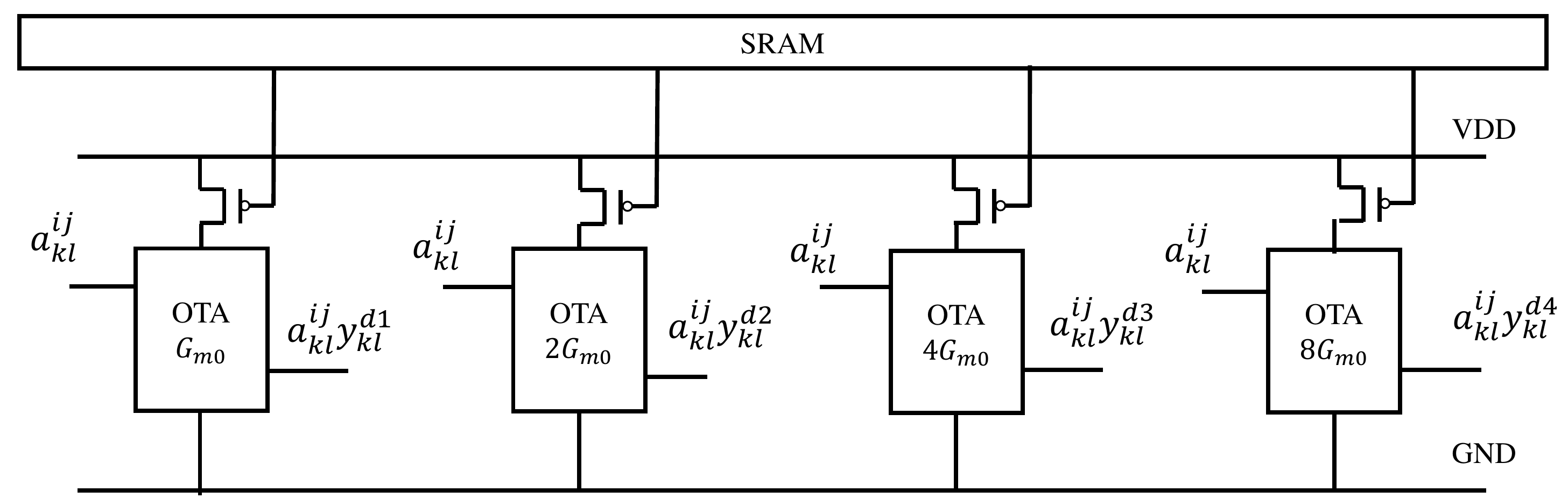}
    \vspace{-8pt}
    \caption{The schematic for an OTA within each CeNN cell for representing 4-bit weights. Data from SRAM connects to the transistor switches to power gate OTAs, so as to program the CeNN template to different values.}
    \label{fig:programmable_schematic}
    \vspace{-0.15in}
\end{figure}

Many prior works have focused on CeNNs implemented by analog circuits using CMOS transistors. 
Per Sec. \ref{sec:2}, a widely used implementation is based on OTAs \cite{Chou97_short}. Here, an OTA is built with two-stage operational amplifiers \cite{Qiuwen15_short}. We use $N$ OTAs with quantized $g_m$ values (i.e., $g_{m0}$, $2g_{m0}$, ..., $2^{N-1}g_{m0}$) to realize N-bit templates (i.e., weights).
The $g_{m0}$'s values are set according to the power requirement since $g_m$'s values are proportional to the bias current. 
Each OTA is connected to a switch for power gating. By power gating different combinations of these OTAs (as shown in Fig. \ref{fig:programmable_schematic}), different template values can be realized.

The cell resistance ($R_{cell}$ in Fig. \ref{fig:CNN}) here is set as $1/g_m$ ($g_m=2^{N}g_{m0}$) such that the cell voltage $x$ settles to the desired output to achieve correct CoNN functionality. The cell capacitance ($C_{cell}$ in Fig. \ref{fig:CNN}) is the summation of the output capacitance of nearby OTAs.
The delay and energy estimation of a CeNN cell in this paper is different from that in \cite{DATE17_short} in that: (1) 32 nm technology is used for the hardware design, (2) the $g_m's$ of the OTAs are larger for faster processing while still satisfying a given power requirement, and (3) the cell resistance $R_{cell}$ in \cite{DATE17_short} is assumed to be the absolute value of the sum of $g_m's$, which leads to much larger settling times. Therefore, the work in \cite{DATE17_short} is a conservative estimation and overestimates the delay and the energy.

\subsection{Analog memory design}
\label{ssec:analog_momory}
In order to support operations that may require multiple (analog) steps associated with different CeNN templates, each CeNN cell is augmented by an embedded analog memory array \cite{analog_memory_short} (see Fig. \ref{fig:architecture}). For the CeNN based convolution computation described in Sec. \ref{ssec:architecture}, analog memory is used to store the intermediate result after each step. For a convolution layer, all the intermediate results described in Algorithm \ref{alg:dataflow} need to be stored in the analog memory. 
The design of the analog memory and the op amp are from \cite{analog_memory_short}.
Specifically, the analog memory array is implemented by a write transistor ($T_w$) and read transistor ($T_r$) to enable write and read. An additional op amp is used to hold the state of the capacitors shown in Fig.~\ref{fig:architecture}b. Multiple pass transistors and capacitors $C_{mem}$ are used to store data. 
Each capacitor $C_{mem}$ and pass transistor forms a memory cell (as a charge storage capacitor) within the analog memory array that could store one state value of a CeNN cell (i.e., data correspond to one pixel). The number of capacitors ($C_{mem}$) within one analog memory array depends on the data needed to store in the memory. The gates of the pass transistors are connected to a MUX. Thus,
$V_{s1}$ to $V_{sx}$ shown in Fig. \ref{fig:architecture}b are controlled by the MUX to determine which capacitor memory needs to be written/read. If the analog signal needs to write to the memory, transistor $T_w$ is on, and one of the pass transistors is selected by the MUX. The data is written to the corresponding capacitor $C_{mem}$. For a read, transistor $T_r$ is on, and one of the pass transistors is selected by MUX.
As each analog memory array is dedicated to one CeNN cell, CeNN cells can access these memory arrays in parallel. 

\vspace{-5pt}

\subsection{SRAM}
\label{sec:5:sram}
An SRAM array is used to store all the template values required for CeNNs to realize a CoNN. While the SRAM itself is a standard design, we still need to carefully select the number of bitlines within one word line due to power and performance constraints.
One design choice may have one word containing all the template values for one CeNN array. For one template operation, $10N_b$ bits are needed for $N_b$-bit precision weight (including 9 template values and a bias). For this option, if $N$ CeNN arrays have distinct sets of templates (i.e., in the convolution layer), $N$ accesses will be required. However, if $N$ CeNN arrays have the same templates (i.e., in the ReLU and pooling layers), only 1 access is required. To reduce the number of accesses, two or more $10N_b$-bit words may be accessed in one cycle by either using more read ports or longer SRAM words. 
After SRAM cell data are read, they are used to control how an OTA is power gated, which in turn realizes different weight values.

\subsection{ADC and hardware for FC layers}
\label{ssec:fc_layer}
Each CeNN is connected to an ADC to convert analog data to a digital representation whenever necessary, e.g., for FC layer computations (i.e., the last layer in Fig. \ref{fig:design1} computation).
FC layers typically require computing the dot product of two large vectors. Such operations are not well-suited for CeNNs with limited kernel size. Hence, either a CPU, GPU, or other hardware should be employed. In the benchmarking efforts to be discussed in Sec. \ref{sec:6}, combinations of digital adders, multipliers, and registers (i.e., ASICs) are used. For simplicity, ripple carry adders and array multipliers are employed in our simulations. Both inputs and weights are $N_b$ bits (where $N_b$ refers the to the precision of CeNN). We also assume that the weights for the FC layer are stored in SRAM. The result of the multiplication is $2N_b$ bits, while an additional $N_b$ bits are used to store the final results of this layer to avoid overflow. Thus, there are $3N_b$ bits at the output. That said, alternative network designs as shown in Sec. \ref{sec:4:EliminatingFCLayer} can eliminate this layer.

\section{Evaluation}
\label{sec:6}
We now evaluate the architectures, networks, and algorithms described above to determine {\bf (i)} whether or not CeNN-friendly CoNNs are competitive with respect to existing architectures and algorithms that address the same dataset and {\bf (ii)} if so, what elements of the CeNN design space lead to superior application-level FOM (e.g., energy and delay per classification, and accuracy). While our CeNN architecture can be applied to different datasets, we specifically compare our approach to other efforts in the context of the MNIST and CIFAR-10 dataset given the wealth of comparison points available.

\subsection{Simulation setup}
\label{sec:6:setup}
Components of the CeNN-based architecture are evaluated via SPICE simulation using the Arizona State University Predictive Technology Model (ASU PTM) for high-performance MOSFET devices at the 32 nm technology node \cite{zhao06_short}.
We use CACTI 7 \cite{cacti_short} to estimate the delay and energy needed for SRAM accesses with the same technology node. The size of SRAM is set as 16 KB to retain reasonable access time/energy, while also accommodating all templates for the proposed networks. The SRAM can be scaled if necessary, to accommodate all the weights in larger networks. In our SRAM design, each wordline contains $10N$ bitlines, so that all weights needed for one CeNN operation can be read from SRAM only once. The analog memory is also scaled to the same technology node.

Though the architecture itself can realize any number of bits, we assume 4-bit and 8-bit precision in our evaluation. 4-bit results help to inform the energy efficiency of our design with reasonable application-level classification accuracy, while 8-bit designs generally do not sacrifice accuracy when compared with 32-bit floating point representation. 
We use 4 CeNNs that correspond to 4 feature maps in the networks described in Sec. \ref{sec:4} for evaluation. However, the number of CeNNs could be changed as a tradeoff between processing time and area/power, as discussed in Algorithm \ref{alg:dataflow} in Sec. \ref{ssec:architecture}.
We take the trained model from TensorFlow, and perform inference computations in a MATLAB based infrastructure with both feature maps and weights quantized to 4 bits or 8 bits to predict accuracy. 


The supply voltage is set to 1 V, and the ratio of the current mirrors in the OTAs is set to 2, to save power in the first stage of OTA. For different precision requirements, the same OTA schematic is used with different transistor sizes and bias currents.  The multiple OTA design in Sec. \ref{ssec:cell} could be used to represent different number of bits for weights. These OTAs are reprogrammed in each step. Here, for each OTA, we convert the signal-to-noise ratio (SNR) of OTA to bit precision using the methods in \cite{snr_bits} to represent different number of bits for feature maps. Compared to the 4-bit designs, the 8-bit designs increase the bias current by 7.5$\times$, and increase the transistor width by 4$\times$ to increase the SNR of the circuit from 32.1 dB to 50.6 dB. Thus, the delay increases by 4.3$\times$ due to the change of bias conditions and increase of transistor size (i.e., parasitic capacitance increases), and the power increases by 7.5$\times$ as the bias current increases. The $g_m$'s of an OTA can be selected to tradeoff processing speed and power. Here, we use four OTAs with $g_m$ values 12$uA/V$, 24$uA/V$, 48$uA/V$, and 96$uA/V$, to realize 4-bit templates (i.e., weights). In the 8-bit design, larger granularity is used, and $g_m$ values are set to 0.75$uA/V$, 1.5$uA/V$, 3$uA/V$, 6$uA/V$, 12$uA/V$, 24$uA/V$, 48$uA/V$, and 96$uA/V$.
We assume state-of-the-art ADC designs \cite{ADC16_short, ADC14_short} to estimate the delay and energy of analog to digital conversion needed before the FC layer in the network in Fig. \ref{fig:design1}. We assume each CeNN is associated with an ADC to convert analog data to digital representation. 

We employ the same device model to benchmark analog memory arrays. We first determine the capacitance and size of pass transistors based on the methods in \cite{analog_memory_short}. The capacitance is $C_{mem} = 55fF$ and the width of the transistor is 180 nm. We use a minimum length of 30 nm. Then, memory write time is determined by the resistance of pass transistor $T_{pass}$ and the capacitor $C_{mem}$. The memory read time is determined by the analog signal through the read buffer. We use SPICE to measure the delay of the analog memories. Per simulations, each memory write and read requires 124 $ps$ and 253 $ps$, respectively.

To satisfy the precision requirements, we also study the robustness of our architecture by evaluating the PVT condition with four corner cases (FF 80\textdegree{}C, SS -40\textdegree{}C, FS 27\textdegree{}C, SF 27\textdegree{}C). We also apply a 5\% variation on the supply voltage to study the impact to the OTA in the CeNN cell, which is the essential computational element in our design. 
Since the $g_m$ of the OTAs in CeNN cell represents the template values in the CeNN operations, we evaluated the $g_m$ variations in the OTA design in these corner cases in the PVT condition study. We specifically focus on the OTA with the largest $g_m$ value in our design, since the variation of that OTA will have the largest impact on the multiplication results. Our simulation results show that in the worst corner case, the error of the circuit still satisfies the precision requirements.
Regarding parasitic capacitance, we have not yet completed a layout of the architecture and cannot precisely model the impact of parasitic capacitance. However, (1) parasitic capacitances within a cell are smaller than the cell capacitance $C_{cell}$ shown in Fig. \ref{fig:architecture}(b), and (2) we assume a CeNN has only local connections with radius of 1 (i.e., to implement 3$\times$3 kernels). Thus, we expect the interconnect parasitic capacitance to be small as a given cell is only connected to its immediate neighbors.

\subsection{Evaluation of the CeNN based architecture}
\label{sec:6:evaluation}
We initially use the 4-bit CeNN design as an example to show how we evaluate the accuracy, delay and energy of our CeNN architecture for performing CoNN computations. We use MNIST as the benchmarking dataset, and both network in Fig. \ref{fig:design1} and network in Fig. \ref{fig:design2} with different configurations (summarized in Table \ref{tbl:mnist_accuracy}) are used for evaluation. 8-bit results are also presented here.

We first measure the energy and delay associated with each layer of a CeNN-friendly CoNN for the 4-bit design. Table \ref{tbl:energy_delay} summaries the delay and energy for each layer, for the networks in Fig. \ref{fig:design1} and Fig. \ref{fig:design2}.
Per Table \ref{tbl:energy_delay}, the energy for each layer in the network in Fig. \ref{fig:design2} decreases with subsequent layers as data is down-sampled, and only a subset of cells in a CeNN are used for the computation. However, delay remains constant (for each layer) as all computations in CeNN cells occur in parallel. (The network in Fig. \ref{fig:design2} has a higher latency than the network in Fig. \ref{fig:design1} in the CeNN components due to the fact that more layers are employed to properly downscale the image, i.e., more template operations are required.) We use the MATLAB framework to quantize the weights and inputs to 4 bits in the inference stage, and classification accuracies for each design are shown in Table \ref{tbl:summary}.
We find that for all cases, the accuracy decreases about 2\% for each design compared with the 32-bit floating point design shown in Table \ref{tbl:mnist_accuracy}, due to the reduced precision of input and weights for our simple network.

We next consider the impact of the ADCs and the FC layer. The delay and energy for an ADC can be approximated based on a 28 nm SAR ADCs design from \cite{ADC16_short}. The total time and energy to port all analog data to the digital domain for the network in Fig. \ref{fig:design1} are 166.7 $ns$ and 3834 $pJ$, respectively (using time multiplexing). For the FC layer, we first use the uniform beyond-CMOS benchmarking (BCB) methodology \cite{2015_Nikonov_short} to estimate the delay and energy for a full adder as well as the register for storing temporary data during the computation.
Then, we estimate the delay of multiplication and addition operations by counting the number of full adders in the critical path of the multiplier and adder. The energy per operation is estimated by the summation of all full adder operations and loading/storing data during computation. The energy and delay overhead due to the interconnect parasitics is also taken into account by using the BCB methodology. Overall, the delay and energy of the FC layer are 124.4 $ns$ and 4041 $pJ$, and they contribute 23\% and 20\% to the total delay and energy per classification for the network  in Fig. \ref{fig:design1} (including ADCs), respectively.

\begin{table}
\vspace{-0.1in}
 \caption{Delay and energy for each CeNN layer}
 \vspace{-0.1in}
 \centering
 \scriptsize
\renewcommand{\arraystretch}{1.35}
 \begin{tabular}{| c " c |c " c|c| }
 \hline
 &\multicolumn{2}{c"}{Network in Fig. \ref{fig:design1}} &\multicolumn{2}{c|}{Network in Fig. \ref{fig:design2}}  \\
 \hline
   Layer & Delay($ns$) & Energy($pJ$) & Delay($ns$) & Energy($pJ$)\\
 \hline
   Conv. 1 & 5.3 & 626 & 5.3 & 626 \\
 \hline
   ReLU1 & 10.7 & 536 & 10.7 & 536 \\
 \hline
   Pooling1 & 85.5 & 4290 & 85.5 & 3398 \\
 \hline
   Conv. 2 & 42.8 & 2827 & 42.8 & 981 \\
 \hline
   ReLU2 & 10.7 & 410 & 10.7 & 186 \\
 \hline
   Pooling2 & 85.5 & 3277 & 85.5 & 1489\\
 \hline
   Conv. 3 & - & - & 42.8 & 519\\
 \hline
   ReLU3 &-  & - & 10.7 & 115\\
 \hline
   Pooling3 & - & - & 85.5 & 921\\
  \hline
   Conv. 4 & - & - & 53.4 & 582\\
 \hline
   ADC + FC & 291.1 & 7875 & - & -\\
  \hline
  Total & 531.6 & 19841 & 432.9 & 9353 \\
  \hline
  \end{tabular}
  \label{tbl:energy_delay}
  \vspace{-0.1in}
\end{table}

Though the network in Fig. \ref{fig:design2} (with no FC layer) requires additional layers to properly downscale the image, the delay is still 19\% lower than the network in Fig. \ref{fig:design1}. Additionally, the network in Fig. ~\ref{fig:design2} requires 2.1$\times$ less energy per classification due to downsampling. However, the accuracy for the network in Fig. \ref{fig:design2} is 0.5\% lower than that in Fig. \ref{fig:design1}.

To evaluate the impact of different approaches for pooling operations, as well as how non-linear template operations impact energy, delay, and accuracy, we apply each design alternative to the networks in Figs. \ref{fig:design1} and \ref{fig:design2}. Results are summarized in Table \ref{tbl:summary}. The numbers in parenthesis refer to the comparison between the alternative approach with the baseline (i.e., the network in Figs. \ref{fig:design1} and \ref{fig:design2} with maximum pooling and linear templates). By using average pooling, the delay/energy is reduced by 1.4$\times$/1.5$\times$ and 2.2$\times$/2.1$\times$ for the networks in Figs. \ref{fig:design1} and \ref{fig:design2}, respectively -- as 16 CeNN steps are reduced to 1 step. The accuracy is reduced by 0.8\% for the network in Fig. \ref{fig:design1} and 1.7\% for the network in Fig. \ref{fig:design2}, respectively. Designs with non-linear templates lead to reductions in delay/energy of 1.5$\times$/1.7$\times$ and 3.7$\times$/2.8$\times$ for the networks in Figs. \ref{fig:design1} and \ref{fig:design2}, respectively -- as both ReLU and pooling operations are reduced to a single step. However, the accuracy drops by 3.6\% and 4.5\%, respectively, following the same trend as the floating point precision.

\begin{table}[!b]
  \vspace{-0.22in}
 \caption{Accuracy, delay and energy with 4-bit CeNN architecture design}
  \vspace{-0.1in}
 \centering
 \scriptsize
\renewcommand{\arraystretch}{1.35}
 \begin{tabular}{| c " c |c|c"c|c|c| }
 \hline
 &\multicolumn{3}{c"}{Network in Fig. \ref{fig:design1}} &\multicolumn{3}{c|}{Network in Fig. \ref{fig:design2}}  \\
 \hline
   Approach & Accuracy & Delay & Energy & Accuracy & Delay & Energy\\
  \hline
  Baseline & 96.5\% & 532$ns$ & 19.8$nJ$ & 96.0\% & 433$ns$ & 9.4$nJ$\\
 \hline
   Average & 95.7\% & 372$ns$ & 12.5$nJ$ & 94.3\% & 192$ns$  & 4.4$nJ$\\
   Pooling &  &(1.4x) & (1.5x) &  & (2.2x) & (2.1x)\\
 \hline
   Nonlinear & 92.9\% & 357$ns$ & 12.0$nJ$ & 91.5\% & 116$ns$ & 3.4$nJ$\\
   operation & & (1.5x) & (1.7x) & & (3.7x) &(2.8x)\\
  \hline
  \end{tabular}
  \label{tbl:summary}
  \vspace{-0.1in}
\end{table}



It is obvious that the accuracy drops for 4-bit designs (in Table \ref{tbl:summary}) compared with 32-bit floating point designs (in Table \ref{tbl:mnist_accuracy}). Meanwhile, there is evidence that the 8-bit precision for many networks usually do not sacrifice accuracy compared with 32 bit floating point design, and are widely used in the state-of-the-art training and inference engine \cite{Google_TPU_short}. Therefore, we also evaluate accuracy, delay and energy for our 8-bit CeNN design using the same method above to show the tradeoffs. In this design, we use OTAs with an SNR equivalent to 8-bit precision. The weights are also set to 8 bits. We use a different design \cite{adc_2017} to evaluate ADC overhead to reflect converting analog signals to 8-bit digital signals. The inputs and weights of the digital FC layer are also set to 8 bits. The results are summarized in Table \ref{tbl:summary_8bit}. As expected, the delay and energy both increase compared to the 4-bit design by 2.0 - 4.2$\times$ and 3.8 - 7.5$\times$ depending on the specific designs, but the accuracy approaches that of 32 bit floating point data. In this design, the delay and energy of network in Fig. \ref{fig:design2} increase more than that of the network in Fig. \ref{fig:design1}. The computations of the network in Fig. \ref{fig:design2} is mostly in the analog domain, while the computations in the network in Fig. \ref{fig:design1} use both analog and digital circuits. As the number of bit increases, the delay and energy for computations associated with analog circuits increase generally faster than the delay and energy for computations associated with digital circuits.

\begin{table}
 \vspace{-0.1in}
 \caption{Accuracy, delay and energy with 8-bit CeNN architecture design}
  \vspace{-0.1in}
 \centering
 \scriptsize
\renewcommand{\arraystretch}{1.35}
 \begin{tabular}{| c |c |c|c|c|c|c| }
 \hline
 &\multicolumn{3}{|c|}{Network in Fig. \ref{fig:design1}} &\multicolumn{3}{|c|}{Network in Fig. \ref{fig:design2}}  \\
 \hline
   Approach & Accuracy & Delay & Energy & Accuracy & Delay & Energy\\
  \hline
  Baseline & 98.0\% & 1442$ns$ & 104.9$nJ$ & 97.8\% & 1828$ns$ & 56.6$nJ$\\
 \hline
   Average pooling & 97.5\% & 773$ns$ & 49.9$nJ$ & 97.4\% & 819$ns$  & 23.0$nJ$\\
 \hline
   Nonlinear operation & 95.4\% & 710$ns$ & 46.2$nJ$ & 94.2\% & 490$ns$ & 23.6$nJ$\\
  \hline
  \end{tabular}
  \label{tbl:summary_8bit}
  \vspace{-0.1in}
\end{table}

\subsection{Comparison to other MNIST implementations}
\label{compare_to_mnist}
\begin{figure}[!b]
    \centering
    \vspace{-0.1in}
    \includegraphics[width=5in]{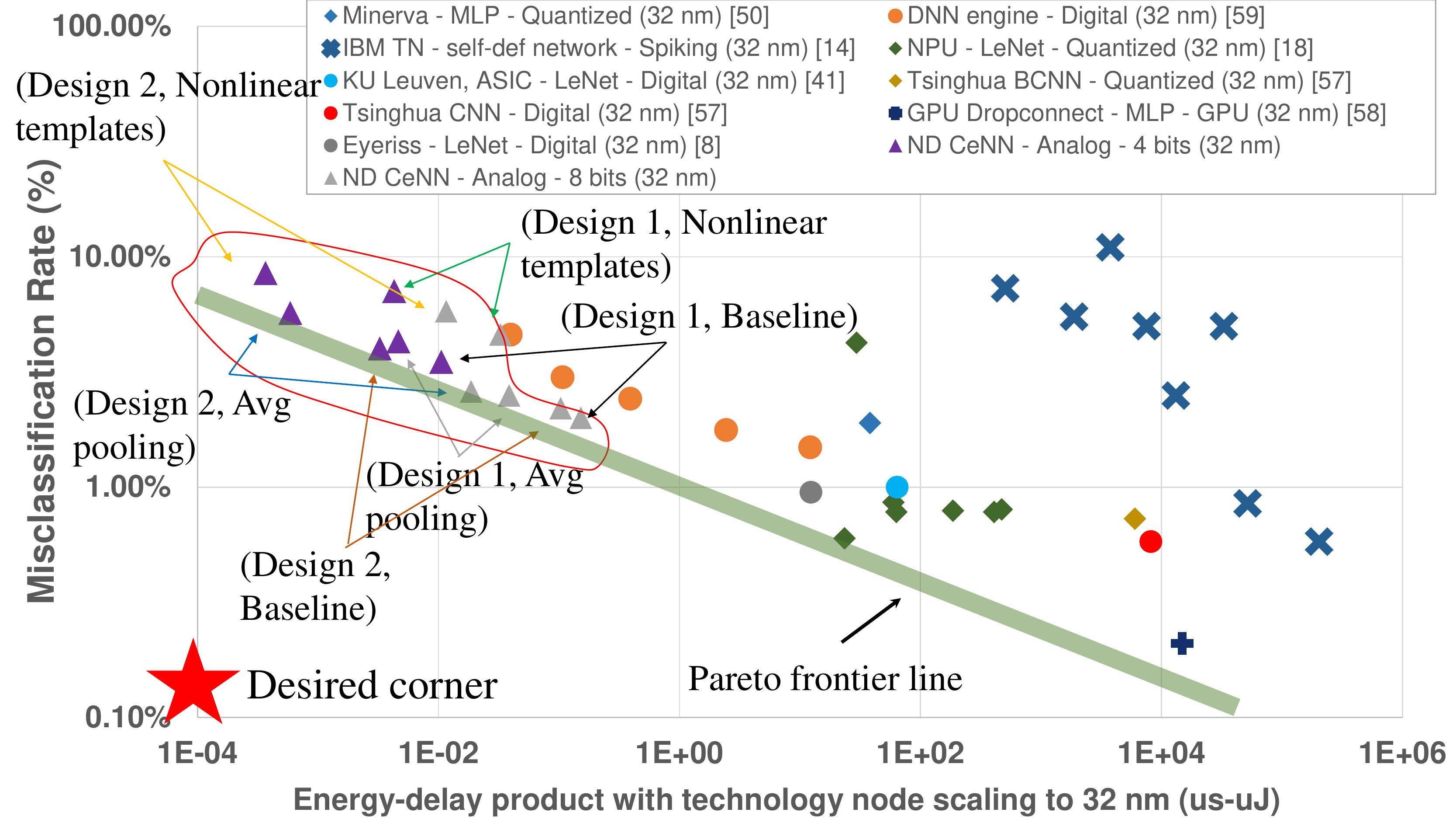}
    \vspace{-0.15in}
    \caption{Benchmarking results for CeNN-friendly CoNNs as well as other algorithms and architectures for the MNIST digit classification problem.}
    \label{fig:benchmarking_scale}
    \vspace{-0.2in}
\end{figure}

It now begs the question as to how our CeNN-based approach compares to other accelerator architectures and algorithms that have been developed to address classification problems such as MNIST. 
Since the computations in our designs are mostly performed in analog domain, we first compare our work with a recent logic-in-memory analog implementation that addresses the same problem \cite{analog_dnn_short}. We compare the delay and energy of convolution layers here. As \cite{analog_dnn_short} only reports the throughput and energy efficiency for the first two convolutional layers in LeNet-5, using 7-bit inputs and 1-bit weights, we also use the throughput and energy efficiency for convolution layers in our baseline network design for fair comparison. The comparison results are shown in Table \ref{tbl:compare_with_analog}. Our CeNN design demonstrates 10.3$\times$ EDP improvements than \cite{analog_dnn_short}. At the application-level, we still obtain better classification accuracy (96.5\% v.s. 96\%). However, since \cite{analog_dnn_short} does not include the data for FC layer, they do not have the complete EDP data on MNIST. Hence, we do not include the implementation in \cite{analog_dnn_short} the benchmarking plot (Fig. \ref{fig:benchmarking_scale}) to be discussed.


\begin{table*}[!b]
 \vspace{-0.1in}
 \caption{Detailed comparison to analog implementation \cite{analog_dnn_short} for MNIST dataset}
  \vspace{-0.1in}
 \centering
 \scriptsize
\renewcommand{\arraystretch}{1.35}
 \begin{tabular}{| c |c |c|c|c|c|c| }
 \hline
   Approach & Precision of & Precision of & Efficiency & Energy Efficiency & Technology & Accuracy \\
   & feature maps & weights & & & & \\
  \hline
  CeNN based approach & 4 bits & 4 bits & 251 GOPS & 12.3 TOPS/W & 32 nm & 96.5\% \\
  \hline
  Logic-in-memory analog circuit \cite{analog_dnn_short} & 7 bits & 1 bit & 10.7 GOPS & 28.1 TOPS/W & 65 nm & 96\% \\
  \hline
  \end{tabular}
  \label{tbl:compare_with_analog}
\end{table*}

We next consider a state-of-the-art digital DNN engine presented in \cite{2017_ISSCC_Whatmough_short} with 28 nm technology node for the MNIST dataset at {\bf iso-accuracy} with our CeNN based design.
We scale the design in \cite{2017_ISSCC_Whatmough_short} from 28 nm to 32 nm for a fair comparison using the method described in \cite{scaling_short}. 
The work in \cite{2017_ISSCC_Whatmough_short} assumes a multilayer perception (MLP) network with 8-bit feature maps and weights, varying the different network sizes. 
Among these different networks, we find three implementations that match the accuracy of our three designs.
Their network sizes are 784$\times$16$\times$16$\times$16$\times$10,  784$\times$32$\times$32$\times$32$\times$10, and 784$\times$64$\times$64$\times$64$\times$10, with accuracy of 95.41\%, 97.0\%, and 97.58\%, respectively.
Meanwhile, our three designs are (i) network in Fig. \ref{fig:design2}, baseline with 4-bit precision (accuracy to be 96.03\%), (ii) network in Fig. \ref{fig:design1}, baseline with 4-bit precision, (96.5\% accuracy) and (iii) network in Fig. \ref{fig:design2}, average pooling with 8-bit precision (97.41\% accuracy). We compare FOMs including energy and delay at iso-accuracy for these designs.

\begin{table*}
 \caption{Detailed comparison to DNN engine \cite{2017_ISSCC_Whatmough_short} for MNIST dataset}
  \vspace{-0.1in}
 \centering
 \scriptsize
\renewcommand{\arraystretch}{1.35}
 \begin{tabular}{| c |c |c|c|c|c|c| }
 \hline
   Comparison & Approach & Accuracy & Bits & Delay (ns) & Energy (nJ) & EDP (nJ-ns) \\
  \Xhline{3\arrayrulewidth}
  \multirow{2}*{Comparison 1} & CeNN -- design 2, baseline & 96.03\% & 4 & 372 & 9.0  & $4.6\times 10^3$ \\
  \cline{2-7}
  & DNN engine \cite{2017_ISSCC_Whatmough_short} & 95.41\% & 8 & 1001 & 39.9 & $4.0\times 10^4$ \\
  \Xhline{3\arrayrulewidth}
  \multirow{2}*{Comparison 2} & CeNN -- design 1, baseline & 96.5\% & 4 & 532 & 19.8 & $1.1\times 10^4$ \\
  \cline{2-7}
  & DNN engine \cite{2017_ISSCC_Whatmough_short} & 97.0\%  & 8 & 1478 & 72.5 & $1.0\times 10^5$ \\
  \Xhline{3\arrayrulewidth}
  \multirow{2}*{Comparison 3} & CeNN -- design 2, avg. pooling & 97.80\%  & 8 & 810 & 230 & $1.9\times 10^5$ \\
  \cline{2-7}
  & DNN engine \cite{2017_ISSCC_Whatmough_short} & 97.58\%  & 8 & 2692 & 145 & $3.9\times 10^5$ \\
  \hline
  \end{tabular}
  \label{tbl:compare_with_dnn}
  \vspace{-0.05in}
\end{table*}


From Table \ref{tbl:compare_with_dnn}, we can find that in our implementation, the EDP and energy efficiency are 2.1 - 8.7$\times$ and 6 - 27$\times$ better, respectively, than the DNN engine \cite{2017_ISSCC_Whatmough_short}. The 8-bit CeNN based design is not as efficient as the 4-bit design with respect to energy efficiency -- compared with the DNN engine due to the fact that analog circuits have worse area/delay/energy compared with digital circuits in higher precision. Here, our delay and energy data is based on simulations, while the data for DNN engine is based on the measurement. Therefore, some discrepancy may exist. However, in general,
with the CeNN approach, (i) high parallelism can be achieved in terms of multiplications and additions in the CeNN-based architecture, (ii) the network exploits local analog memory for fast processing, and (iii) accessing feature maps in the analog domain is faster than accessing the digital weights in the digital domain. Thus, the weight stationary approach is used. That said, once the weights are read from the SRAM (i.e., all the cells are configured), all the computations associated with the weights are performed. The weights do not need to be read from SRAM again. Therefore, the total weight access time is minimized. Since there are still unused OTAs in our design, it may be further optimized to reduce the delay and energy.

We also compare our work with a wider range of implementations, including custom ASIC chips \cite{Reagen16_short, 2017_ISSCC_Whatmough_short, Moons17_short,eyeriss_short}, neural processing units \cite{Hashemi17_short}, spiking neural networks \cite{2015_NIPS_Esser_short, 2015_VLSI_Kim_short,sparse_spiking_short}, crossbar implementations \cite{tsinghua_crossbar_short}, and CPU/GPU-based solutions of the DropConnect approach \cite{Wan15_short} (the most accurate approach for MNIST to date; data is measured via i7-5820K, 32GB DDR3 with Nvidia Titan). 
Fig. \ref{fig:benchmarking_scale} plots the EDP v.s. misclassification rate for all these approaches. In order to make a fair comparison, we again scale all delay/energy data to the 32 nm technology node using the ITRS data based on \cite{scaling_short}. 

Note that the comparison is shown in the log scale, additional uncertainties (interconnects parasitics, clocking, control circuits) should not change the overall trend shown in Fig. \ref{fig:benchmarking_scale} as the EDP of these elements would not be orders of magnitude larger \cite{2015_Nikonov_short}.  Our approach has significantly lower EDP compared with other approaches with comparable classification accuracy. Among our designs, higher EDPs are generally correlated with higher accuracy. We draw a Pareto frontier line (the green line in Fig. \ref{fig:benchmarking_scale} according to the product of misclassification rate and the EDP. In our designs, several datapoints are on the Pareto frontier. Specifically, for the 4-bit design, the network in Fig. \ref{fig:design2} with maximum pooling and linear templates, and the network in Fig. \ref{fig:design2} with average pooling and linear templates are on the Pareto frontier, while for the 8-bit design, the network in Fig. \ref{fig:design1} with average pooling linear templates are on the Pareto frontier in the plot. We should add that the EDP values of some of the implementations \cite{Reagen16_short,2017_ISSCC_Whatmough_short,eyeriss_short,Moons17_short} in Fig. \ref{fig:benchmarking_scale} are obtained from actual measurements, while others are from simulation. Therefore, some discrepancy may exist.

\subsection{Evaluation of larger networks}
In Sec. \ref{compare_to_mnist}, we discussed a comprehensive comparison using the MNIST problem as the context. However, networks for MNIST are relatively simple. In this subsection, we also compare our CeNN design with other implementations that target larger networks, i.e., we compare with
other accelerators that solve the CIFAR-10 problem.

For the CIFAR-10 dataset, images with size 32$\times$32 are used. we also use CeNNs with the same size to enable parallel processing. The evaluation setup is the same as in Sec. \ref{sec:6:setup}. We use the networks discussed in Sec. \ref{sec:4:cifar10}, and summarize our results in Table \ref{tbl:cifar_accuracy_delay_energy}. Here, we use 4-bit design to maximize the energy efficiency, and the accuracy is close to 32 floating point accuracy (given in Table \ref{tbl:cifar_accuracy}).

\begin{table}
 \caption{Accuracy, delay and energy for different noise levels}
  \vspace{-0.1in}
 \centering
 \scriptsize
\renewcommand{\arraystretch}{1.35}
 \begin{tabular}{| c |c |c |c| }
 \hline
   Approach & CeNN-friendly AlexNet & CeNN-friendly AlexNet & CeNN-friendly AlexNet\\
    & C96-C256-C384-C384-C256 & C64-C128-C256-C256-C128 & C64-C128-C128-C128-C64\\
  \hline
  Accuracy & 83.9\% & 82.2\% & 80.8\%\\
  \hline
  Delay ($\mu s$) & 311 & 106 & 47\\
  \hline
  Energy ($\mu J$)& 497 & 169 & 75\\
  \hline
  \end{tabular}
  \label{tbl:cifar_accuracy_delay_energy}
\end{table}

\begin{figure}
    \centering
    \vspace{-0.1in}
    \includegraphics[width=4.5in]{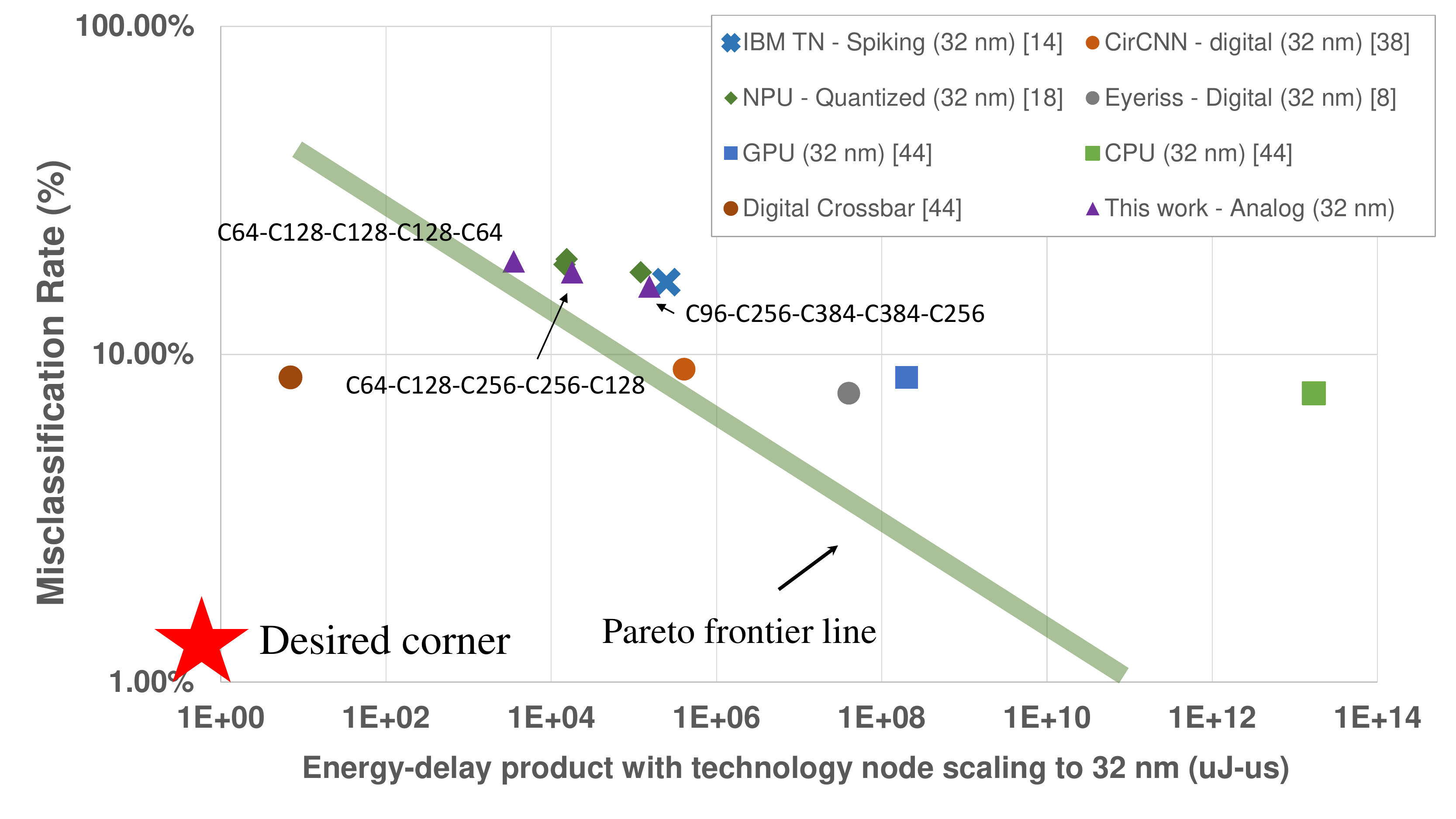}
    \caption{Benchmarking results for CeNN-friendly CoNNs as well as other algorithms and architectures for the CIFAR-10 classification problem.}
    \label{fig:cifar}
    \vspace{-0.2in}
\end{figure}

We compare our approach with a large number of implementations available that solve the CIFAR-10 problem. The benchmarking plot is shown in Fig. \ref{fig:cifar}. The implementation include IBM TrueNorth \cite{2015_NIPS_Esser_short}, Fourier transform approach \cite{fft_cnn}, NPU \cite{Hashemi17_short}, Eyeriss \cite{eyeriss_short}, a mixed-signal approach \cite{mixed_signal_cifar} the CPU and GPU data reported in \cite{digital_crossbar}. We also draw a Pareto frontier line based on the product of misclassification rate and EDP of the data points collected in Fig. \ref{fig:cifar}. From the plot, one of our CeNN datapoint (C64-C128-C128-C128-C64) lands on the Pareto frontier.

We also make an {\bf iso-accuracy} comparison with the NPU data point shown in the plot. We selected a datapoint from our design with similar accuracy to the design in NPU. The detailed comparison is shown in Table \ref{tbl:cifar_comparison}. Not only our accuracy of our CeNN design is 0.3\% better than NPU approach, but also our design achieves 4.3$\times$ EDP compared with the NPU approach. Note that the NPU data are also simulation results.

\begin{table*}[!b]
 \vspace{-0.22in}
 \caption{Detailed comparison to NPU \cite{Hashemi17_short} for the CIFAR-10 dataset}
 \vspace{-0.1in}
 \centering
 \scriptsize
\renewcommand{\arraystretch}{1.35}
 \begin{tabular}{| c |c |c|c|c|c|c|c| }
 \hline
   Approach & Technology node & Accuracy & Bits & Delay ($\mu s$) & Energy ($\mu J$) & EDP ($\mu J$-$\mu s$) \\
  \hline
  CeNN based approach & 32 nm & 80.8\% & 4 & 47 & 75  & 3525 \\
  \hline
  NPU \cite{Hashemi17_short} & 32 nm & 80.5\% & 8 & 485 & 32 & 15332 \\
  \hline
  \end{tabular}
  \label{tbl:cifar_comparison}
  \vspace{-0.1in}
\end{table*}

To articulate our evaluation, we also discuss the differences between our work and other analog accelerators -- i.e., ISAAC \cite{isaac} and RedEye \cite{redeye} here. Our work differs from ISAAC and RedEye in the following aspects.
\begin{enumerate}
    \item Different computation elements are used. ISAAC uses a crossbar architecture, where multiplication and summation are carried out via analog voltage, conductance and current, and signals are accumulated horizontally in the crossbar rows within the chip. RedEye uses tunable capacitors as computation units. Our approach uses CeNN cells as the base element, where multiplications and partial sum calculations are performed using OTAs within each CeNN cell. 
    \item Different dataflows are used. ISAAC uses an in-memory computation architecture, where memristors are used for both storing the weights and performing computation. In RedEye, column-based computation elements are used, and data is passed vertically. In our CeNN architecture, the memory and the computation units are separated. OTAs are used for multiplication while analog memories are used to store intermediate results.
    \item The requirements on devices are different. In ISAAC, memristors are required to perform the logic-in-memory computation. In RedEye, conventional CMOS is used for benchmarking. While we also assume conventional CMOS for benchmarking activities in this paper, our work is compatible with other emerging devices as well (e.g., many emerging devices have been considered in the context of CeNN implementations per \cite{cenn_emerging_devices}). 
\end{enumerate}

\vspace{-0.15in}
\subsection{Training with actual I-V characteristics}
\begin{figure}[!b]
    \centering
    \vspace{-0.15in}
    \includegraphics[width=2in]{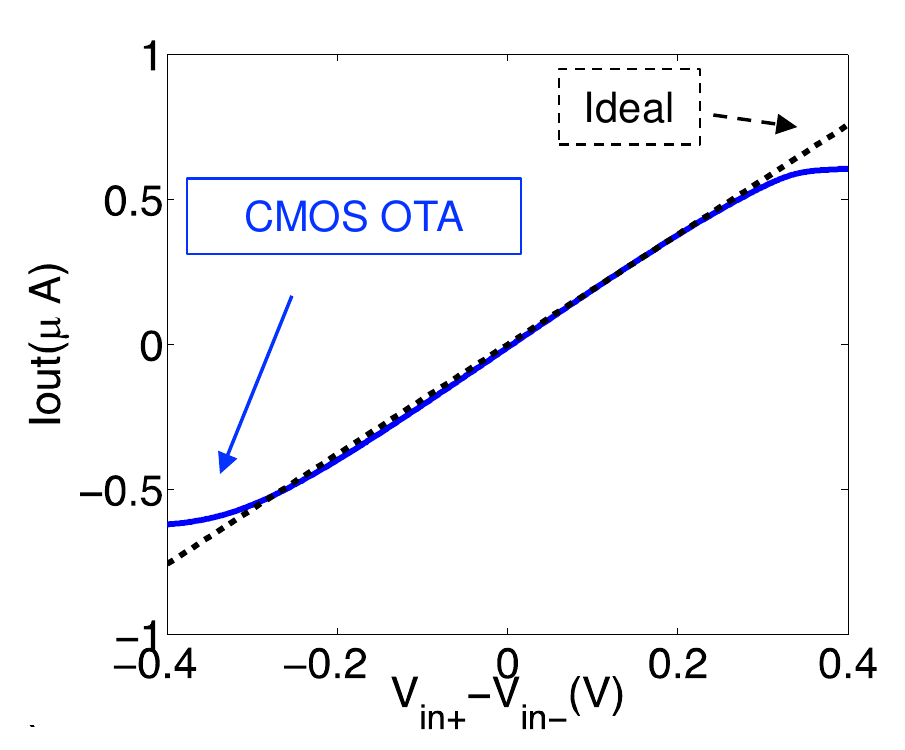}
    \vspace{-0.1in}
    \caption{Ideal I-V characteristics and actual characteristics for OTA design.}
    \label{fig:IV_characteristic}
    \vspace{-0.15in}
\end{figure}

In Sec. \ref{sec:6:evaluation}, we show that by leveraging the 8-bit representation, the accuracy does not decrease much compared with the 32-bit floating point representation. However, another source of error comes from the actual I-V characteristics of an OTA. For example, in Fig. \ref{fig:IV_characteristic}, when the difference of two inputs, $(V_{in+} - V_{in-})$, of the OTA is larger than $0.2V$, the mismatch between the actual and ideal I-V characteristics becomes more severe. This behavior could potentially decrease the accuracy. 

To study the impact, we include the mismatch described above into the inference stage. We use the actual I-V characteristics of an OTA obtained from SPICE simulation, and build a look-up table.
We then embed the table into the MATLAB based CeNN simulator in the inference stage. That is, whenever an OTA operation is required, results for the OTA are read from the lookup table, instead of by direct matrix multiplication. 
Simulations of the networks in Figs. \ref{fig:design1} and \ref{fig:design2} suggest that by including the actual I-V characteristics in the network, the accuracy decreases from 98.1\% and 96.5\% to 96.8\% and 95.8\%, respectively.

However, this accuracy decrease can be largely compensated by leveraging the I-V characteristics in the training stage. We use the same look-up table, and plug it into the forward path of the training stage of the network in the TensorFlow framework. By considering the I-V characteristics during training, the accuracy increases and become close to the ideal accuracy. The results are summarized in Table \ref{tbl:mnist_training}. We can see that by using the actual I-V characteristics in the training stage, the accuracy only decreases 0.2\% when compared with the original network for the baseline design for network in Fig. \ref{fig:design1} and network in Fig. \ref{fig:design2}. This approach should be applicable for other non-ideal circuit behaviors.

\begin{table*}
 \caption{Classification accuracy for MNIST when actual I-V characteristics are included in training/inference}
 \vspace{-0.1in}
 \centering
 \scriptsize
\renewcommand{\arraystretch}{1.35}
 \begin{tabular}{|c|c|c|c|}
 \hline
   Network type & Original network & Inference with actual I-V & Training \& inference \\
   &&& with actual I-V\\
  \hline
   Network in Fig. \ref{fig:design1}, linear templates, max pooling & 98.1\% & 96.5\% & 97.9\% \\
  \hline
   Network in Fig. \ref{fig:design2}, linear templates, max pooling & 97.8\% & 95.8\% & 97.6\% \\
  \hline
  \end{tabular}
  \label{tbl:mnist_training}
  \vspace{-0.2in}
\end{table*}

Whether individualized training might be needed is still an open question. However, existing literature suggests that some PVT variations and noise in the circuit may not greatly impact application level accuracy for both MOSFETs and emerging devices (e.g. see \cite{redeye,device_variation_1}), thus individualized training would not be needed. Researchers have also investigated on-chip training given device variations (e.g., see \cite{on_chip_training_1,on_chip_training_2}), and reasonable application level accuracy results are indeed obtained. Essentially, at present, there are no firm conclusions about whether individualized training will be required. We will also study this in our future work. 

\section{Conclusions and Discussions}
\label{sec:7}
This paper presents a mixed-signal architecture for hardware implementation of convolutional neural networks. The architecture is based on an analog CeNN realization. We demonstrate the use of CeNN to realize different layers of CoNN, and the design of CeNN-friendly CoNNs. We present tradeoffs for each CeNN based design, and compare our approaches with various other existing accelerators to illustrate the benefits for the MNIST and CIFAR-10 problem as case studies. 
Our results show that the CeNN based approach can lead to superior performance while retaining reasonable accuracy. Specifically, 8.7$\times$ EDP for MNIST problem, and 4.3$\times$ EDP for CIFAR-10 problem are obtained in iso-accuracy comparison, when comparing with state-of-the-art approaches. 

Our architecture targets were originally/primarily for edge devices. Network sizes for edge devices (e.g., MobileNet \cite{mobilenet}, SqueezeNet \cite{squeezenet}, etc.) are usually much smaller than AlexNet. Thus, AlexNet for Cifar-10 dataset should be sufficient to illustrate how our approach can be applied to larger networks and how our approach compares other existing works. Furthermore, these networks also only have kernel sizes 3x3 or 1x1, which are suitable for our CeNN computations. We expect that the network model deployed in edge devices should be smaller than our CeNN friendly AlexNet. Thus, our CeNN architecture should be able to process all tasks that could be reasonably processed by IoT devices efficiently. As future work, we will study other larger network topologies to further ensure that reasonable classification accuracies could be obtained (i.e., when compared to published work), and will also consider the CeNN approach with respect to metrics such as energy and delay in the context of these networks.








\begin{acks}
This work was supported in part by the Center for Low Energy Systems
Technology (LEAST), one of six centers of STARnet, a Semiconductor
Research Corporation program sponsored by MARCO and DARPA.
This project was supported by the National Science Foundation under grant 1640081, and the Nanoelectronics Research Corporation (NERC), a wholly-owned subsidiary of the Semiconductor Research Corporation (SRC), through Extremely Energy Efficient Collective Electronics (EXCEL), an SRC-NRI Nanoelectronics Research Initiative under Research Task ID 2698.004.
\end{acks}

\bibliographystyle{ACM-Reference-Format}
\bibliography{sample-bibliography}

\end{document}